\documentclass{article}

\usepackage[
	activate={true,nocompatibility},
	tracking=true
]{microtype}
\usepackage{graphicx}
\usepackage{subfigure}
\usepackage{booktabs} 
\usepackage{amsmath}
\usepackage{mathtools}
\usepackage{makecell}
\usepackage{multirow}
\usepackage{comment}

\newcommand{\modelname}{GLIDE}
\newcommand{\email}[1]{\href{mailto:#1}{#1}}

\makeatletter
\def\blfootnote{\xdef\@thefnmark{}\@footnotetext}
\makeatother

\usepackage[accepted]{icml2021}

\usepackage{hyperref}
\usepackage[all]{hypcap}

\icmltitlerunning{\modelname{}: Towards Photorealistic Image Generation and Editing with Text-Guided Diffusion Models}

\begin{document}

\twocolumn[
\icmltitle{\modelname{}: Towards Photorealistic Image Generation and Editing with Text-Guided Diffusion Models}



\icmlsetsymbol{equal}{*}

\begin{icmlauthorlist}
\icmlauthor{Alex Nichol}{equal}
\icmlauthor{Prafulla Dhariwal}{equal}
\icmlauthor{Aditya Ramesh}{equal}
\icmlauthor{Pranav Shyam}{}
\icmlauthor{Pamela Mishkin}{}
\icmlauthor{Bob McGrew}{}
\icmlauthor{Ilya Sutskever}{}
\icmlauthor{Mark Chen}{}
\end{icmlauthorlist}

\icmlcorrespondingauthor{Alex Nichol}{alex@openai.com}
\icmlcorrespondingauthor{Prafulla Dhariwal}{prafulla@openai.com}
\icmlcorrespondingauthor{Aditya Ramesh}{aramesh@openai.com}

\icmlkeywords{Machine Learning, image generation, image editing, image inpainting, diffusion models, guided diffusion, CLIP, contrastive learning}

\vskip 0.3in
]




\begin{abstract}
Diffusion models have recently been shown to generate high-quality synthetic images, especially when paired with a guidance technique to trade off diversity for fidelity. We explore diffusion models for the problem of text-conditional image synthesis and compare two different guidance strategies: CLIP guidance and classifier-free guidance. We find that the latter is preferred by human evaluators for both photorealism and caption similarity, and often produces photorealistic samples. Samples from a 3.5~billion parameter text-conditional diffusion model using classifier-free guidance are favored by human evaluators to those from DALL-E, even when the latter uses expensive CLIP reranking. Additionally, we find that our models can be fine-tuned to perform image inpainting, enabling powerful text-driven image editing. We train a smaller model on a filtered dataset and release the code and weights at \href{https://github.com/openai/glide-text2im}{https://github.com/openai/glide-text2im}.
\end{abstract}

\section{Introduction}
\label{sec:introduction}

\blfootnote{$^*$Equal contribution. Correspondence to \email{alex@openai.com}, \email{prafulla@openai.com}, \email{aramesh@openai.com}}

\begin{figure*}[h!]
    \centering
    \setlength{\tabcolsep}{2.0pt}
    \begin{tabular}{ccccc}
        \includegraphics[width=0.22\textwidth]{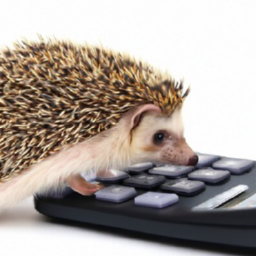} &
        \includegraphics[width=0.22\textwidth]{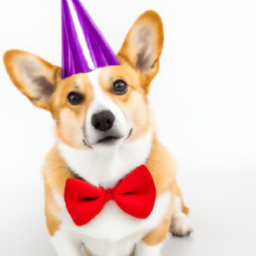} &
        \includegraphics[width=0.22\textwidth]{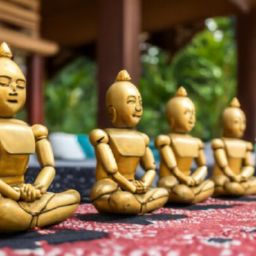} &
        \includegraphics[width=0.22\textwidth]{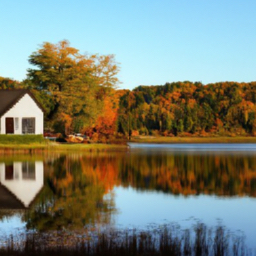} \\

        \scriptsize \makecell{``a hedgehog using a \\ calculator''} &
        \scriptsize \makecell{``a corgi wearing a red bowtie \\ and a purple party hat''} &
        \scriptsize \makecell{``robots meditating in a \\ vipassana retreat''} &
        \scriptsize \makecell{``a fall landscape with a small \\ cottage next to a lake''} \\
        \rule{0pt}{0.2pt} \\
        
        \includegraphics[width=0.22\textwidth]{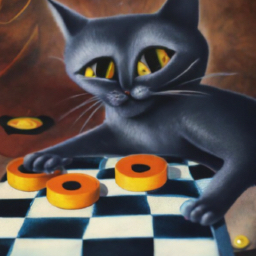} &
        \includegraphics[width=0.22\textwidth]{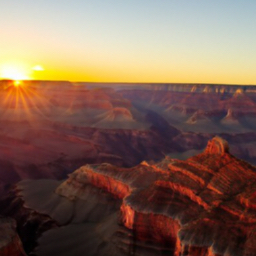} &
        \includegraphics[width=0.22\textwidth]{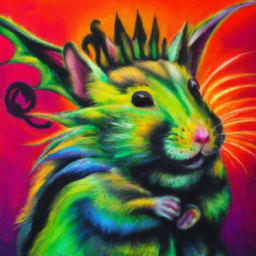} &
        \includegraphics[width=0.22\textwidth]{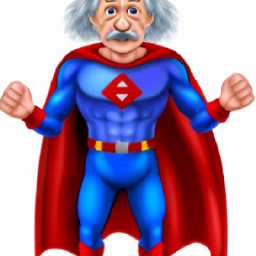} \\
        \scriptsize \makecell{``a surrealist dream-like oil \\ painting by salvador dalí \\ of a cat playing checkers''} &
        \scriptsize \makecell{``a professional photo of a \\ sunset behind the grand \\ canyon''} &
        \scriptsize \makecell{``a high-quality oil painting \\ of a psychedelic hamster \\ dragon''} &
        \scriptsize \makecell{``an illustration of albert \\ einstein wearing a superhero \\ costume''} \\

        \rule{0pt}{0.2pt} \\
        
        \includegraphics[width=0.22\textwidth]{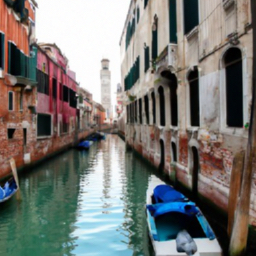} &
        \includegraphics[width=0.22\textwidth]{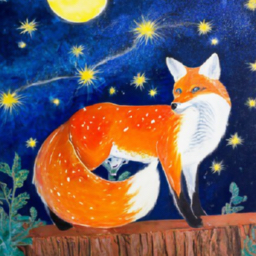} &
        \includegraphics[width=0.22\textwidth]{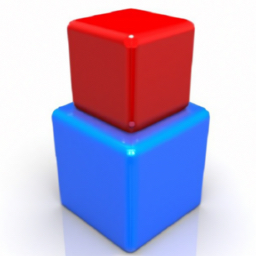} &
        \includegraphics[width=0.22\textwidth]{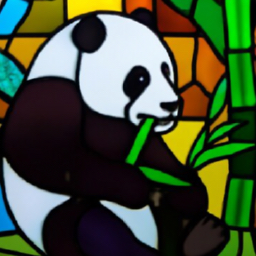} \\

        \scriptsize \makecell{``a boat in the canals of venice''} &
        \scriptsize \makecell{``a painting of a fox in the style \\ of starry night''} &
        \scriptsize \makecell{``a red cube on top \\ of a blue cube''} &
        \scriptsize \makecell{``a stained glass window \\ of a panda eating bamboo''} \\
        
        \rule{0pt}{0.2pt} \\
        \includegraphics[width=0.22\textwidth]{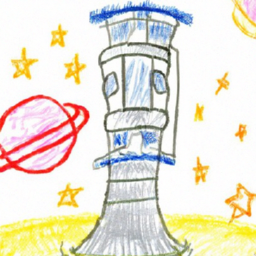} &
        \includegraphics[width=0.22\textwidth]{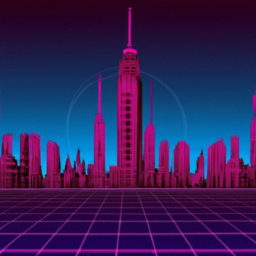} &        
        \includegraphics[width=0.22\textwidth]{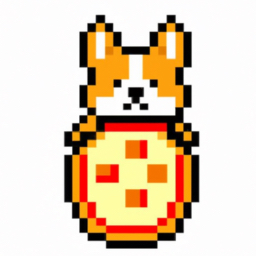} &
        \includegraphics[width=0.22\textwidth]{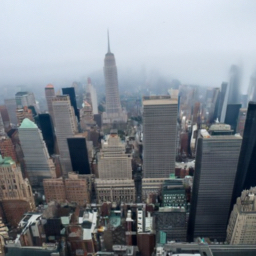} \\
        \scriptsize \makecell{``a crayon drawing of a space elevator''} &
        \scriptsize \makecell{``a futuristic city in synthwave style''} &
        \scriptsize \makecell{``a pixel art corgi pizza''} &
        \scriptsize \makecell{``a fog rolling into new york''} \\

    \end{tabular}

    \caption{Selected samples from \modelname{} using classifier-free guidance. We observe that our model can produce photorealistic images with shadows and reflections, can compose multiple concepts in the correct way, and can produce artistic renderings of novel concepts. For random sample grids, see Figure \ref{fig:guidance_comp_bamboo} and \ref{fig:guidance_comp_living}.}
    \label{fig:header_samples}
    \vskip -0.1in
\end{figure*}

\begin{figure*}[t]
    \centering
    \setlength{\tabcolsep}{4.0pt}
    \begin{tabular}{cc}
        \includegraphics[width=0.45\textwidth]{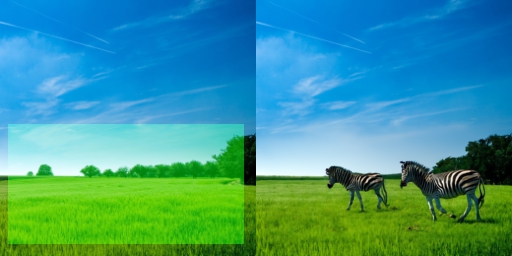} &
        \includegraphics[width=0.45\textwidth]{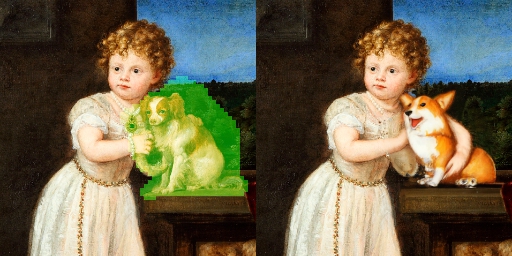} \\

        \scriptsize \makecell{``zebras roaming in the field''} &
        \scriptsize \makecell{``a girl hugging a corgi on a pedestal''} \\

        \rule{0pt}{0.5pt} \\

        \includegraphics[width=0.45\textwidth]{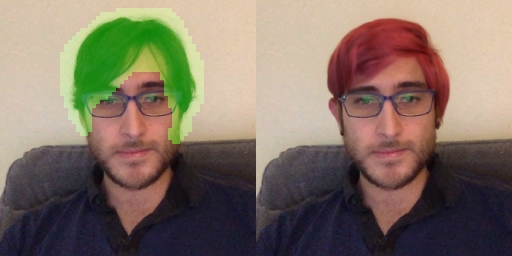} &
        \includegraphics[width=0.45\textwidth]{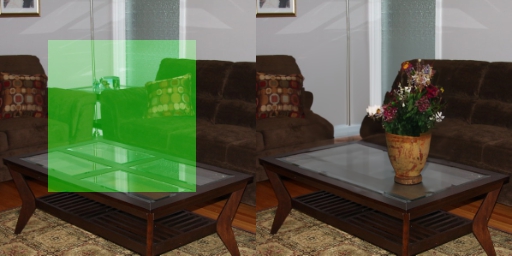} \\

        \scriptsize \makecell{``a man with red hair''} &
        \scriptsize \makecell{``a vase of flowers''} \\
        
        \rule{0pt}{0.5pt} \\

        \includegraphics[width=0.45\textwidth]{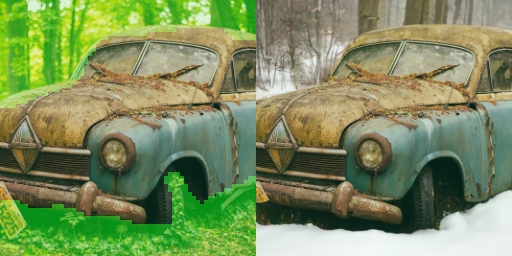} &
        \includegraphics[width=0.45\textwidth]{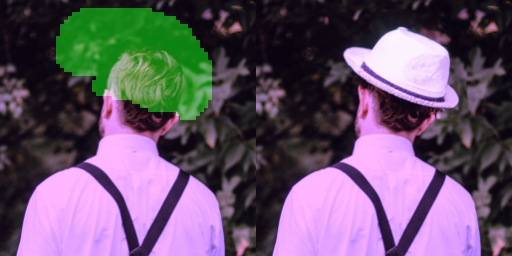} \\

        \scriptsize \makecell{``an old car in a snowy forest''} &
        \scriptsize \makecell{``a man wearing a white hat''}
    \end{tabular}

    \caption{Text-conditional image inpainting examples from \modelname{}. The green region is erased, and the model fills it in conditioned on the given prompt. Our model is able to match the style and lighting of the surrounding context to produce a realistic completion.}
    \label{fig:inpainting_examples}
    \vskip -0.1in 
\end{figure*}

\begin{figure*}[t]
    \centering
    \setlength{\tabcolsep}{1.0pt}
    \begin{tabular}{ccccc}
        \includegraphics[width=0.19\textwidth]{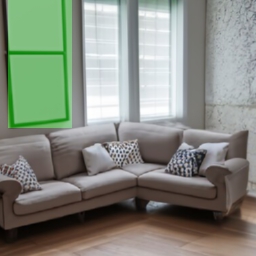} &
        \includegraphics[width=0.19\textwidth]{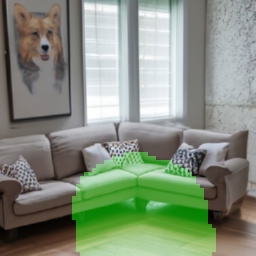} &
        \includegraphics[width=0.19\textwidth]{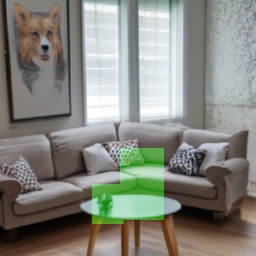} &
        \includegraphics[width=0.19\textwidth]{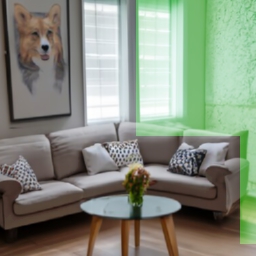} &
        \includegraphics[width=0.19\textwidth]{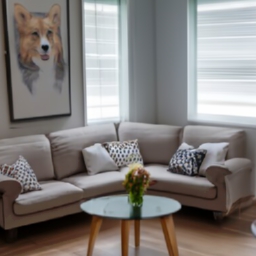} \\

        \scriptsize \makecell{``a cozy living room''} &
        \scriptsize \makecell{``a painting of a corgi \\ on the wall above \\ a couch''} &
        \scriptsize \makecell{``a round coffee table \\ in front of a couch''} &
        \scriptsize \makecell{``a vase of flowers on a \\ coffee table''} &
        \scriptsize \makecell{``a couch in the corner \\ of a room''}
    \end{tabular}

    \caption{Iteratively creating a complex scene using \modelname{}. First, we generate an image for the prompt ``a cozy living room'', then use the shown inpainting masks and follow-up text prompts to add a painting to the wall, a coffee table, and a vase of flowers on the coffee table, and finally to move the wall up to the couch.}
    \label{fig:iterative_inpainting}
    \vskip -0.1in 
\end{figure*}

\begin{figure}[t]
    \centering
    \begin{tabular}{c}
        \includegraphics[width=0.45\textwidth]{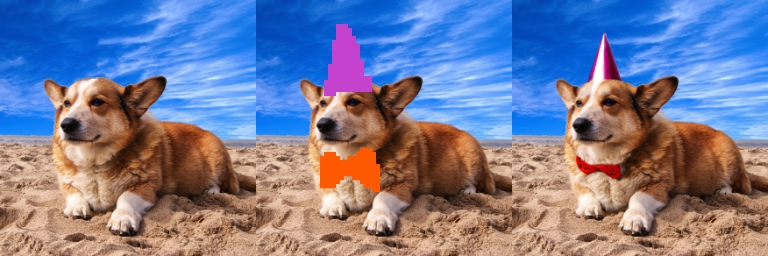} \\
        \scriptsize \makecell{``a corgi wearing a bow tie and a birthday hat''} \\ 
        \rule{0pt}{0.5pt} \\

        \includegraphics[width=0.45\textwidth]{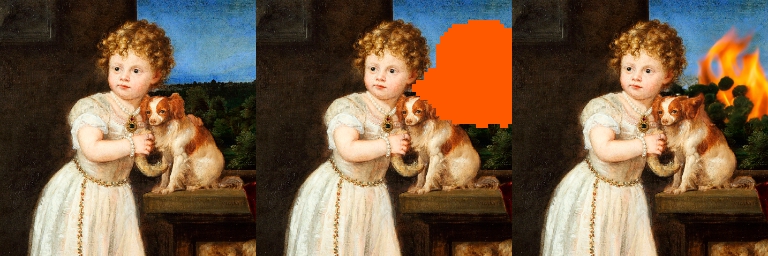} \\
        \scriptsize \makecell{``a fire in the background''} \\ 
        \rule{0pt}{0.5pt} \\

        \includegraphics[width=0.45\textwidth]{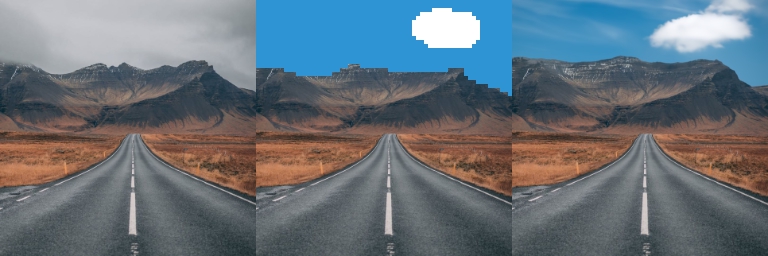} \\
        \scriptsize \makecell{``only one cloud in the sky today''}
    \end{tabular}

    \caption{Examples of text-conditional SDEdit \citep{sdedit} with \modelname{}, where the user combines a sketch with a text caption to do more controlled modifications to an image.}
    \label{fig:sdedit_examples}
    \vskip -0.1in
\end{figure}


Images, such as illustrations, paintings, and photographs, can often be easily described using text, but can require specialized skills and hours of labor to create. Therefore, a tool capable of generating realistic images from natural language can empower humans to create rich and diverse visual content with unprecedented ease. The ability to edit images using natural language further allows for iterative refinement and fine-grained control, both of which are critical for real world applications.

Recent text-conditional image models are capable of synthesizing images from free-form text prompts, and can compose unrelated objects in semantically plausible ways \citep{attngan,dmgan,dfgan,dalle,xmcgan}. However, they are not yet able to generate photorealistic images that capture all aspects of their corresponding text prompts.

On the other hand, unconditional image models can synthesize photorealistic images \cite{biggan,stylegan,stylegan2,vqvae2}, sometimes with enough fidelity that humans can't distinguish them from real images \cite{zhou2019hype}. Within this line of research, diffusion models \cite{dickstein,scorematching} have emerged as a promising family of generative models, achieving state-of-the-art sample quality on a number of image generation benchmarks \citep{ddpm,sotapaper,cascaded}.

To achieve photorealism in the class-conditional setting, \citet{sotapaper} augmented diffusion models with \textit{classifier guidance}, a technique which allows diffusion models to condition on a classifier's labels. The classifier is first trained on noised images, and during the diffusion sampling process, gradients from the classifier are used to guide the sample towards the label. \citet{uncond} achieved similar results without a separately trained classifier through the use of \textit{classifier-free guidance}, a form of guidance that interpolates between predictions from a diffusion model with and without labels.

Motivated by the ability of guided diffusion models to generate photorealistic samples and the ability of text-to-image models to handle free-form prompts, we apply guided diffusion to the problem of text-conditional image synthesis. First, we train a 3.5~billion parameter diffusion model that uses a text encoder to condition on natural language descriptions. Next, we compare two techniques for guiding diffusion models towards text prompts: CLIP guidance and classifier-free guidance. Using human and automated evaluations, we find that classifier-free guidance yields higher-quality images.

We find that samples from our model generated with classifier-free guidance are both photorealistic and reflect a wide breadth of world knowledge. When evaluated by human judges, our samples are preferred to those from DALL-E \citep{dalle} 87\% of the time when evaluated for photorealism, and 69\% of the time when evaluated for caption similarity.  

While our model can render a wide variety of text prompts zero-shot, it can can have difficulty producing realistic images for complex prompts. Therefore, we provide our model with editing capabilities in addition to zero-shot generation, which allows humans to iteratively improve model samples until they match more complex prompts. Specifically, we fine-tune our model to perform image inpainting, finding that it is capable of making realistic edits to existing images using natural language prompts. Edits produced by the model match the style and lighting of the surrounding context, including convincing shadows and reflections. Future applications of these models could potentially aid humans in creating compelling custom images with unprecedented speed and ease.

We observe that our resulting model can significantly reduce the effort required to produce convincing disinformation or Deepfakes. To safeguard against these use cases while aiding future research, we release a smaller diffusion model and a noised CLIP model trained on filtered datasets.

We refer to our system as \modelname{}, which stands for \textbf{G}uided \textbf{L}anguage to \textbf{I}mage \textbf{D}iffusion for Generation and \textbf{E}diting. We refer to our small filtered model as \modelname{} (filtered).

\section{Background}
\label{sec:background}

In the following sections, we outline the components of the final models we will evaluate: diffusion, classifier-free guidance, and CLIP guidance.

\subsection{Diffusion Models}
\label{sec:diffusionbackground}

We consider the Gaussian diffusion models introduced by \citet{dickstein} and improved by \citet{scorematching,ddpm}. Given a sample from the data distribution $x_0 \sim q(x_0)$, we produce a Markov chain of latent variables $x_1, ..., x_T$ by progressively adding Gaussian noise to the sample:
$$q(x_t | x_{t-1}) \coloneqq \mathcal{N}(x_t; \sqrt{\alpha_t} x_{t-1}, (1-\alpha_t)\mathcal{I})$$
If the magnitude $1-\alpha_t$ of the noise added at each step is small enough, the posterior $q(x_{t-1} | x_{t})$ is well-approximated by a diagonal Gaussian. Furthermore, if the magnitude $1 -\alpha_1 ... \alpha_T$ of the total noise added throughout the chain is large enough, $x_T$ is well approximated by $\mathcal{N}(0, \mathcal{I})$. These properties suggest learning a model $p_\theta(x_{t-1} | x_t)$ to approximate the true posterior:
$$p_{\theta}(x_{t-1}|x_t) \coloneqq \mathcal{N}(\mu_{\theta}(x_t), \Sigma_{\theta}(x_t))$$
which can be used to produce samples $x_0 \sim p_\theta(x_0)$ by starting with Gaussian noise $x_T \sim \mathcal{N}(0, \mathcal{I})$ and gradually reducing the noise in a sequence of steps $x_{T-1}, x_{T-2}, ..., x_0$.

While there exists a tractable variational lower-bound on $\log p_\theta(x_0)$, better results arise from optimizing a surrogate objective which re-weighs the terms in the VLB. To compute this surrogate objective, we generate samples $x_t \sim q(x_t | x_0)$ by applying Gaussian noise $\epsilon$ to to $x_0$, then train a model $\epsilon_{\theta}$ to predict the added noise using a standard mean-squared error loss:
$$L_{\text{simple}} \coloneqq E_{t \sim [1,T],x_0 \sim q(x_0), \epsilon \sim \mathcal{N}(0, \mathbf{I})}[||\epsilon - \epsilon_{\theta}(x_t, t)||^2] \label{eq:lsimple}$$

\citet{ddpm} show how to derive $\mu_{\theta}(x_t)$ from $\epsilon_{\theta}(x_t, t)$, and fix $\Sigma_{\theta}$ to a constant. They also show the equivalence to previous denoising score-matching based models \cite{scorematching,improvedscore}, with the score function $\nabla_{x_t} \log p(x_t) \propto \epsilon_{\theta}(x_t, t)$. In a follow-up work, \citet{improved} present a strategy for learning~$\Sigma_{\theta}$, which enables the model to produce high quality samples with fewer diffusion steps. We adopt this technique in training the models in this paper.

Diffusion models have also been successfully applied to image super-resolution \cite{improved,sr3}. Following the standard formulation of diffusion, high-resolution images $y_0$ are progressively noised in a sequence of steps. However, $p_\theta(y_{t-1} | y_t, x)$ additionally conditions on the downsampled input~$x$, which is provided to the model by concatenating $x$ (bicubic upsampled) in the channel dimension. Results from these models outperform prior methods on FID, IS, and in human comparison scores.

\subsection{Guided Diffusion}
\label{sec:guideddiffusion}

\citet{sotapaper} find that samples from class-conditional diffusion models can often be improved with classifier guidance, where a class-conditional diffusion model with mean~$\mu_{\theta}(x_t|y)$ and variance $\Sigma_{\theta}(x_t|y)$ is additively perturbed by the gradient of the log-probability~$\log p_{\phi}(y|x_t)$ of a target class~$y$ predicted by a classifier. The resulting new perturbed mean~$\hat{\mu}_{\theta}(x_t|y)$ is given by 
$$\hat{\mu}_{\theta}(x_t|y) = \mu_{\theta}(x_t|y) + s \cdot \Sigma_{\theta}(x_t|y) \nabla_{x_t} \log p_{\phi}(y|x_t)$$
The coefficient $s$ is called the guidance scale, and \citet{sotapaper} find that increasing $s$ improves sample quality at the cost of diversity.

\subsection{Classifier-free guidance}
\label{sec:uncondguidance}
\citet{uncond} recently proposed classifier-free guidance, a technique for guiding diffusion models that does not require a separate classifier model to be trained. For classifier-free guidance, the label~$y$ in a class-conditional diffusion model~$\epsilon_{\theta}(x_t|y)$ is replaced with a null label $\emptyset$ with a fixed probability during training. During sampling, the output of the model is extrapolated further in the direction of~$\epsilon_{\theta}(x_t|y)$ and away from~$\epsilon_{\theta}(x_t|\emptyset)$ as follows:
$$\hat{\epsilon}_{\theta}(x_t|y) = \epsilon_{\theta}(x_t|\emptyset) + s \cdot (\epsilon_{\theta}(x_t|y) - \epsilon_{\theta}(x_t|\emptyset))$$
Here~$s \ge 1$ is the guidance scale. This functional form is inspired by the implicit classifier $$p^{i}(y|x_t) \propto \frac{p(x_t|y)}{p(x_t)}$$ whose gradient can be written in terms of the true scores~$\epsilon^{*}$ 
\begin{align*}
\nabla_{x_t} \log p^{i}(x_t|y) &\propto \nabla_{x_t} \log p(x_t|y) - \nabla_{x_t} \log p(x_t) \\
&\propto \epsilon^{*}(x_t|y) - \epsilon^{*}(x_t)    
\end{align*}
To implement classifier-free guidance with generic text prompts, we sometimes replace text captions with an empty sequence (which we also refer to as~$\emptyset$) during training. We then guide towards the caption $c$ using the modified prediction~$\hat{\epsilon}$:
$$\hat{\epsilon}_{\theta}(x_t|c) = \epsilon_{\theta}(x_t|\emptyset) + s \cdot (\epsilon_{\theta}(x_t|c) - \epsilon_{\theta}(x_t|\emptyset))$$
Classifier-free guidance has two appealing properties. First, it allows a single model to leverage its own knowledge during guidance, rather than relying on the knowledge of a separate (and sometimes smaller) classification model. Second, it simplifies guidance when conditioning on information that is difficult to predict with a classifier (such as text).

\subsection{CLIP Guidance}
\label{sec:clip}

\citet{clip} introduced CLIP as scalable approach for learning joint representations between text and images. A CLIP model consists of two separate pieces: an image encoder~$f(x)$ and a caption encoder~$g(c)$. During training, batches of~$(x, c)$ pairs are sampled from a large dataset, and the model optimizes a contrastive cross-entropy loss that encourages a high dot-product $f(x) \cdot g(c)$ if the image~$x$ is paired with the given caption $c$, or a low dot-product if the image and caption correspond to different pairs in the training data.



Since CLIP provides a score of how close an image is to a caption, several works have used it to steer generative models like GANs towards a user-defined text caption \cite{clipglass,styleclip,bigsleep,stylegannada}. To apply the same idea to diffusion models, we can replace the classifier with a CLIP model in classifier guidance. In particular, we perturb the reverse-process mean with the gradient of the dot product of the image and caption encodings with respect to the image:
$$\hat{\mu}_{\theta}(x_t|c) = \mu_{\theta}(x_t|c) + s \cdot \Sigma_{\theta}(x_t|c) \nabla_{x_t} \left(f(x_t) \cdot g(c) \right)$$
Similar to classifier guidance, we must train CLIP on noised images $x_t$ to obtain the correct gradient in the reverse process. Throughout our experiments, we use CLIP models that were explicitly trained to be noise-aware, which we refer to as noised CLIP models.

Prior work \citet{clipdiff,secondarymodelmethod} has shown that the public CLIP models, which have not been trained on noised images, can still be used to guide diffusion models. In Appendix~\ref{app:unnoised_comp}, we show that our noised CLIP guidance performs favorably to this approach without requiring additional tricks like data augmentation or perceptual losses. We hypothesize that guiding using the public CLIP model adversely impacts sample quality because the noised intermediate images encountered during sampling are out-of-distribution for the model.

\section{Related Work}
\label{sec:related_work}

Many works have approached the problem of text-conditional image generation. \citet{attngan,dmgan,dfgan,xmcgan,textcl} train GANs with text-conditioning using publicly available image captioning datasets. \citet{dalle} synthesize images conditioned on text by building on the approach of \citet{vqvae}, wherein an autoregressive generative model is trained on top of discrete latent codes. Concurrently with our work, \citet{vqdiff} train text-conditional discrete diffusion models on top of discrete latent codes, finding that the resulting system can produce competitive image samples.

Several works have explored image inpainting with diffusion models. \citet{sdedit} finds that diffusion models can not only inpaint regions of an image, but can do so conditioned on a rough sketch (or set of colors) for the image. \citet{palette} finds that, when trained directly on the inpainting task, diffusion models can smoothly inpaint regions of an image without edge artifacts.

CLIP has previously been used to guide image generation. \citet{clipglass,styleclip,bigsleep,stylegannada} use CLIP to guide GAN generation towards text prompts. The online AI-generated art community has produced promising early results using unnoised CLIP-guided diffusion \citep{clipdiff,secondarymodelmethod}. \citet{diffusionclip} edits images using text prompts by fine-tuning a diffusion model to target a CLIP loss while reconstructing the original image's DDIM \citep{ddim} latent. \mbox{\citet{lafite}} trains GAN models conditioned on perturbed CLIP image embeddings, resulting in a model which can condition images on CLIP text embeddings. None of these works explore noised CLIP models, and often rely on data augmentations and perceptual losses as a result.

Several works have explored text-based image editing. \citet{textinpainting} propose a dual attention mechanism for using text embeddings to inpaint missing regions of an image. \citet{textstylegan} propose a method for editing images of faces using feature vectors grounded in text. \citet{paintbyword} pair CLIP with state-of-the-art GAN models to inpaint images using text targets. Concurrently with our work, \citet{blendeddiff} use CLIP-guided diffusion to inpaint regions of images conditioned on text.

\section{Training}

For our main experiments, we train a 3.5 billion parameter text-conditional diffusion model at $64 \times 64$ resolution, and another 1.5 billion parameter text-conditional upsampling diffusion model to increase the resolution to $256 \times 256$. For CLIP guidance, we also train a noised $64 \times 64$ ViT-L CLIP model \citep{vit}.

\subsection{Text-Conditional Diffusion Models}

We adopt the ADM model architecture proposed by \citet{sotapaper}, but augment it with text conditioning information. For each noised image~$x_t$ and corresponding text caption $c$, our model predicts $p(x_{t-1}|x_t, c)$. To condition on the text, we first encode it into a sequence of~$K$ tokens, and feed these tokens into a Transformer model \citep{transformer}. The output of this transformer is used in two ways: first, the final token embedding is used in place of a class embedding in the ADM model; second, the last layer of token embeddings (a sequence of $K$ feature vectors) is separately projected to the dimensionality of each attention layer throughout the ADM model, and then concatenated to the attention context at each layer.

We train our model on the same dataset as DALL-E \citep{dalle}. We use the same model architecture as the ImageNet $64 \times 64$ model from \citet{sotapaper}, but scale the model width to 512~channels, resulting in roughly 2.3~billion parameters for the visual part of the model. For the text encoding Transformer, we use 24 residual blocks of width 2048, resulting in roughly 1.2~billion parameters.

Additionally, we train a 1.5~billion parameter upsampling diffusion model to go from $64 \times 64$ to $256 \times 256$ resolution. This model is conditioned on text in the same way as the base model, but uses a smaller text encoder with width~1024 instead of~2048. Otherwise, the architecture matches the ImageNet upsampler from \citet{sotapaper}, except that we increase the number of base channels to~384.

We train the base model for 2.5M~iterations at batch size~2048. We train the upsampling model for 1.6M~iterations at batch size~512. We find that these models train stably with 16-bit precision and traditional loss scaling \citep{lossscaling}. The total training compute is roughly equal to that used to train DALL-E.

\subsection{Fine-tuning for classifier-free guidance}

After the initial training run, we fine-tuned our base model to support unconditional image generation. This training procedure is exactly like pre-training, except 20\%~of text token sequences are replaced with the empty sequence. This way, the model retains its ability to generate text-conditional outputs, but can also generate images unconditionally.

\begin{figure*}[h!]
    \centering
    \setlength{\tabcolsep}{2.0pt}
    \begin{tabular}{cccccc}
        \rotatebox{90}{\scriptsize\phantom{AAAA} Real Image} &
        \includegraphics[width=0.18\textwidth]{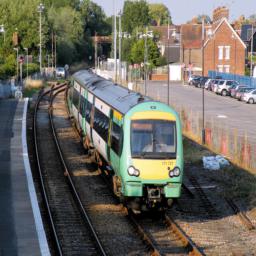} &
        \includegraphics[width=0.18\textwidth]{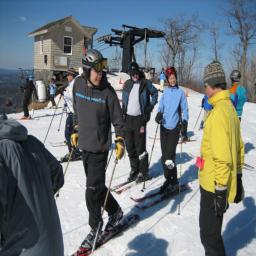} &
        \includegraphics[width=0.18\textwidth]{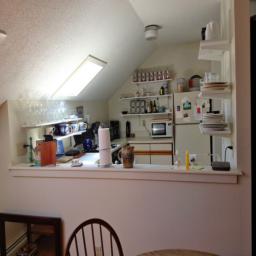} &
        \includegraphics[width=0.18\textwidth]{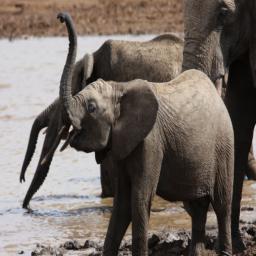} &
        \includegraphics[width=0.18\textwidth]{figures/xmcgan_mscoco/original/4.jpg} \\

        \rotatebox{90}{\scriptsize\phantom{AAAA} XMC-GAN} &
        \includegraphics[width=0.18\textwidth]{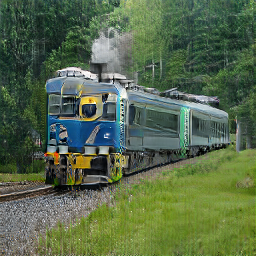} &
        \includegraphics[width=0.18\textwidth]{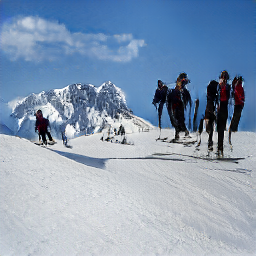} &
        \includegraphics[width=0.18\textwidth]{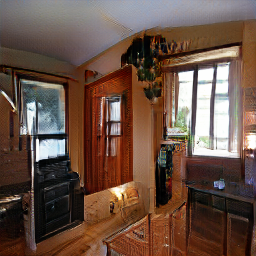} &
        \includegraphics[width=0.18\textwidth]{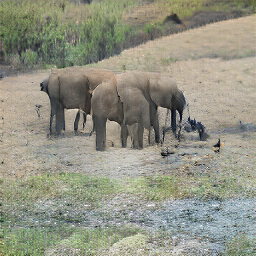} &
        \includegraphics[width=0.18\textwidth]{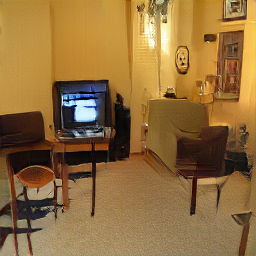} \\
        
        \rotatebox{90}{\scriptsize\phantom{AAAAA} DALL-E} &
        \includegraphics[width=0.18\textwidth]{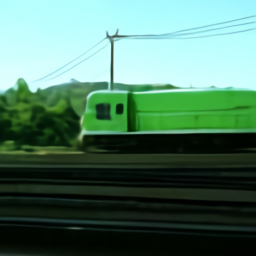} &
        \includegraphics[width=0.18\textwidth]{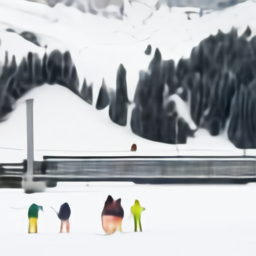} &
        \includegraphics[width=0.18\textwidth]{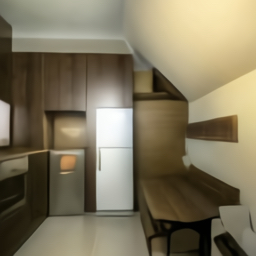} &
        \includegraphics[width=0.18\textwidth]{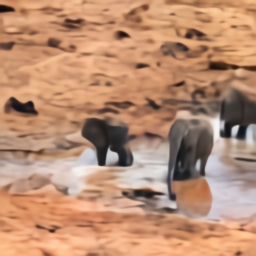} &
        \includegraphics[width=0.18\textwidth]{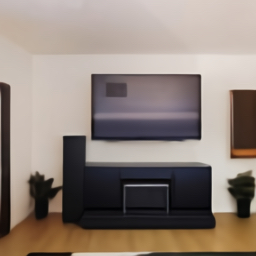} \\

        \rotatebox{90}{\scriptsize\phantom{AAA} \modelname{} (CLIP Guid.)} &
        \includegraphics[width=0.18\textwidth]{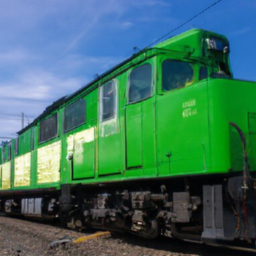} &
        \includegraphics[width=0.18\textwidth]{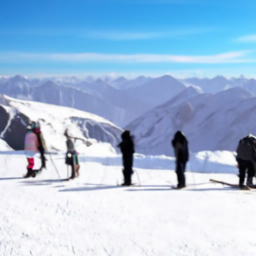} &
        \includegraphics[width=0.18\textwidth]{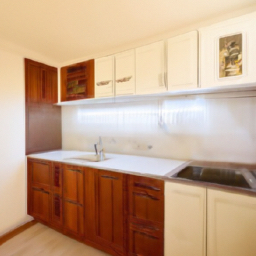} &
        \includegraphics[width=0.18\textwidth]{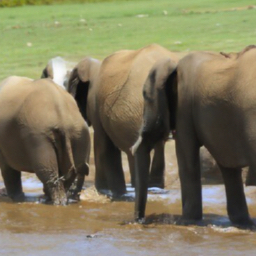} &
        \includegraphics[width=0.18\textwidth]{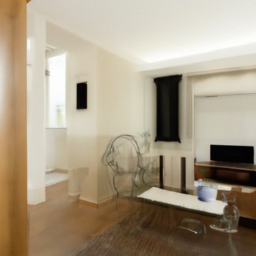} \\

        \rotatebox{90}{\scriptsize\phantom{AA} \modelname{} (CF Guid.)} &
        \includegraphics[width=0.18\textwidth]{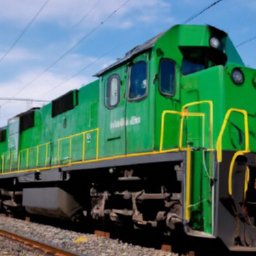} &
        \includegraphics[width=0.18\textwidth]{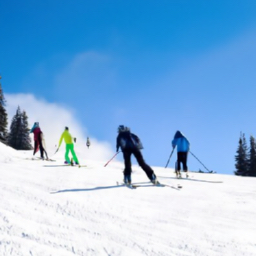} &
        \includegraphics[width=0.18\textwidth]{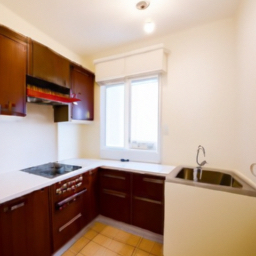} &
        \includegraphics[width=0.18\textwidth]{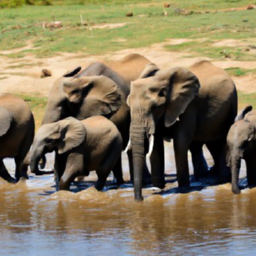} &
        \includegraphics[width=0.18\textwidth]{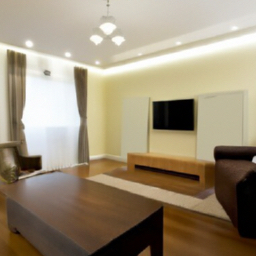} \\

        & \scriptsize \makecell{``a green train is coming \\ down the tracks''}
        & \scriptsize \makecell{``a group of skiers are \\ preparing to ski down \\ a mountain.''}
        & \scriptsize \makecell{``a small kitchen with \\ a low ceiling''}
        & \scriptsize \makecell{``a group of elephants walking \\ in muddy water.''}
        & \scriptsize \makecell{``a living area with a \\ television and a table''}
    \end{tabular}
    \caption{Random image samples on MS-COCO prompts. For XMC-GAN, we take samples from \citet{xmcgan}. For DALL-E, we generate samples at temperature~0.85 and select the best of~256 using CLIP reranking. For \modelname{}, we use CLIP guidance with scale~2.0 and classifier-free guidance with scale~3.0. We do not perform any CLIP reranking or cherry-picking for \modelname{}.}
    \label{fig:coco_model_comparison}
    \vskip -0.1in
\end{figure*}

\begin{figure*}[h!]
    \centering
    \subfigure[Precision/Recall]{
        \includegraphics[width=0.3\textwidth]{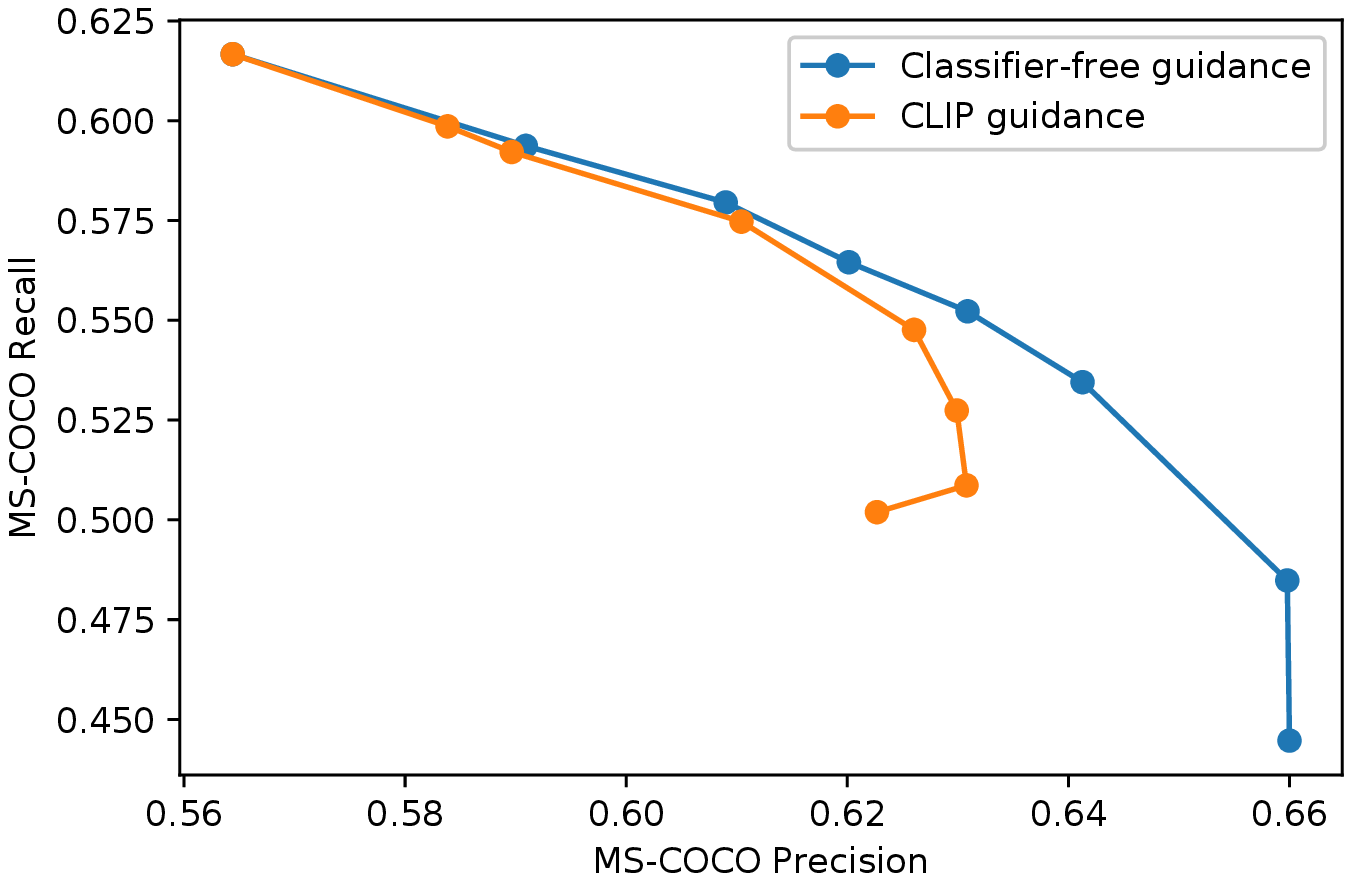}
    }
    \subfigure[IS/FID]{
        \includegraphics[width=0.3\textwidth]{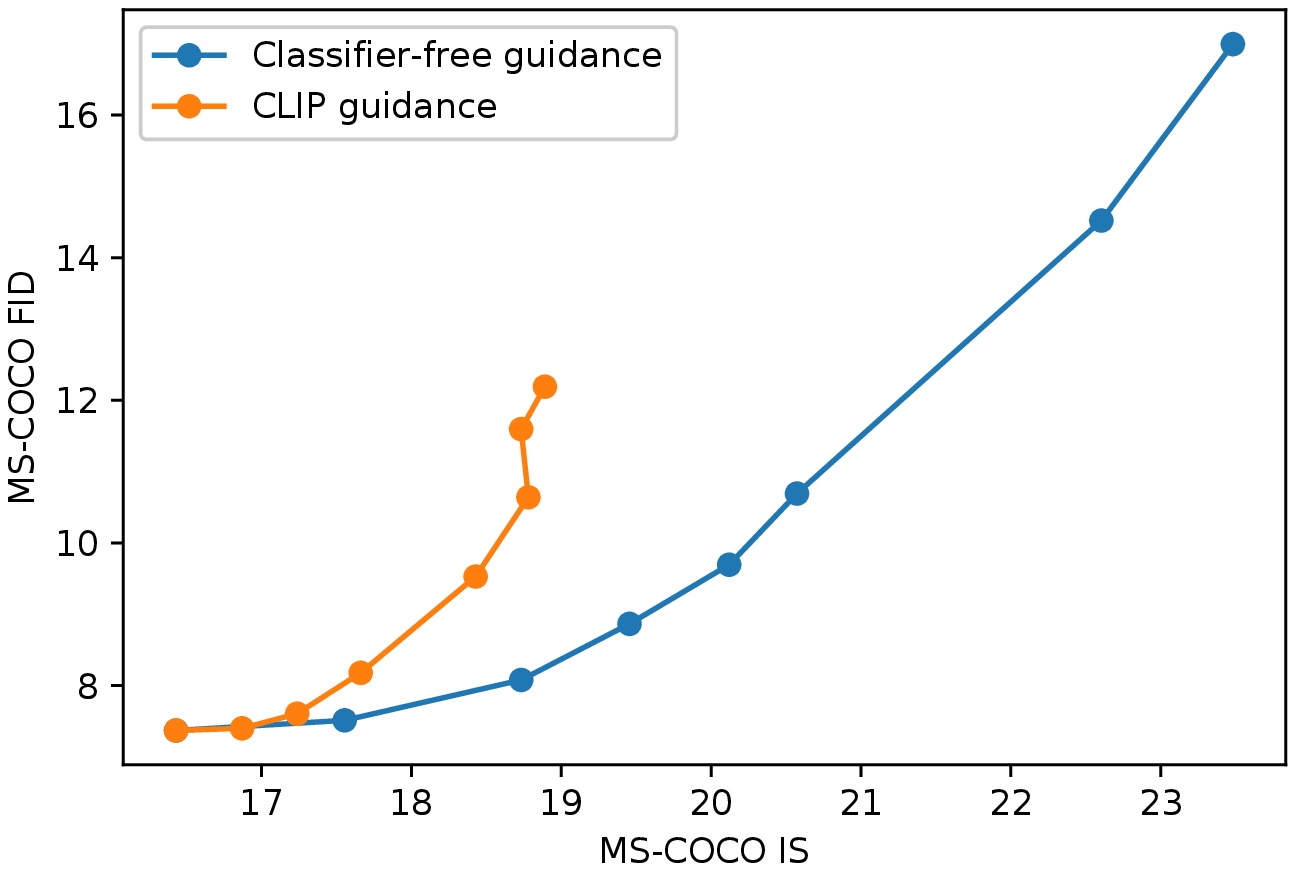}
    }
    \subfigure[CLIP score/FID]{
        \includegraphics[width=0.3\textwidth]{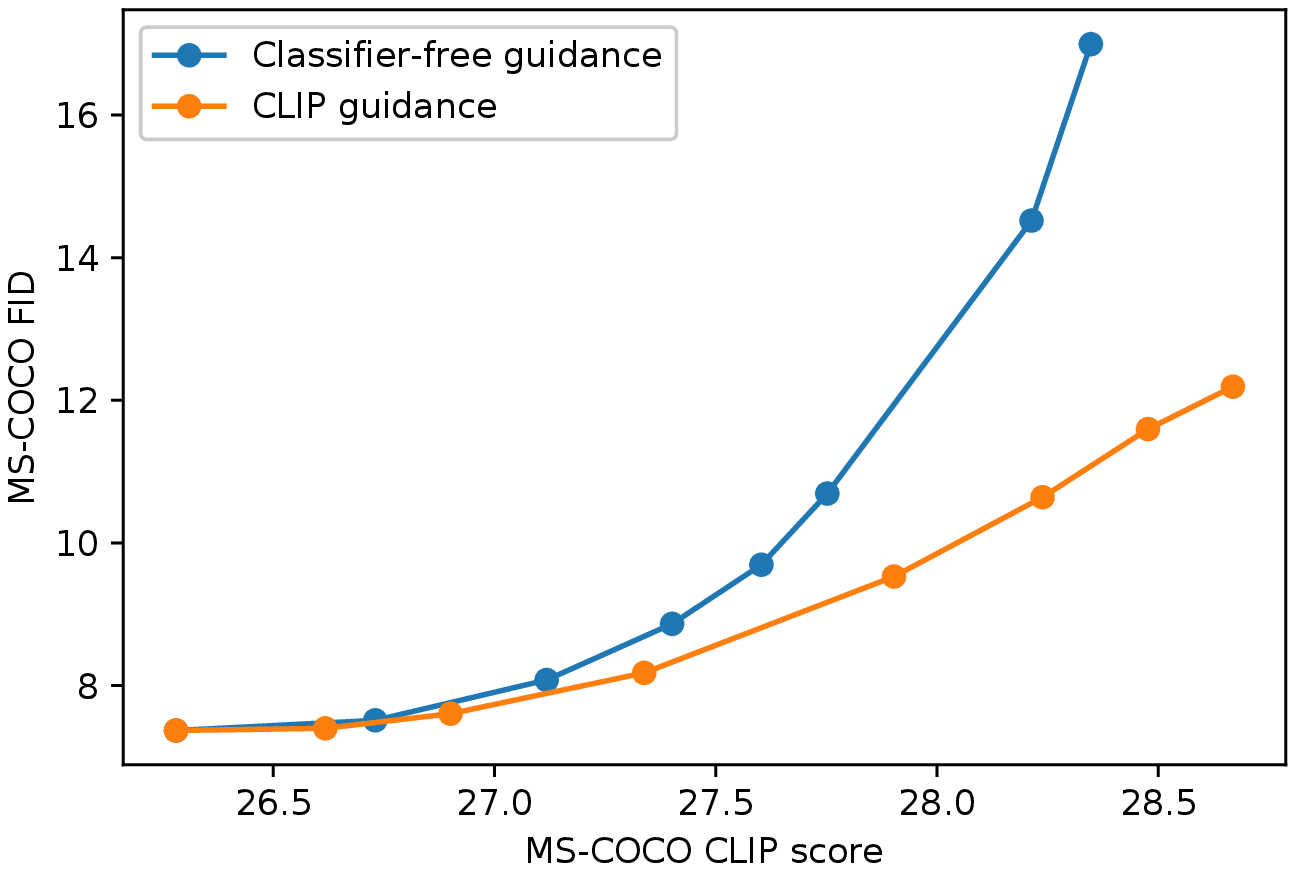}
    }
    \caption{Comparing the diversity-fidelity trade-off of classifier-free guidance and CLIP guidance on MS-COCO $64 \times 64$.}
    \label{fig:tradeoff_evals}
    \vskip -0.1in
\end{figure*}

\begin{figure}[h!]
    \begin{center}
    \subfigure[Photorealism]{
        \includegraphics[width=0.32\textwidth]{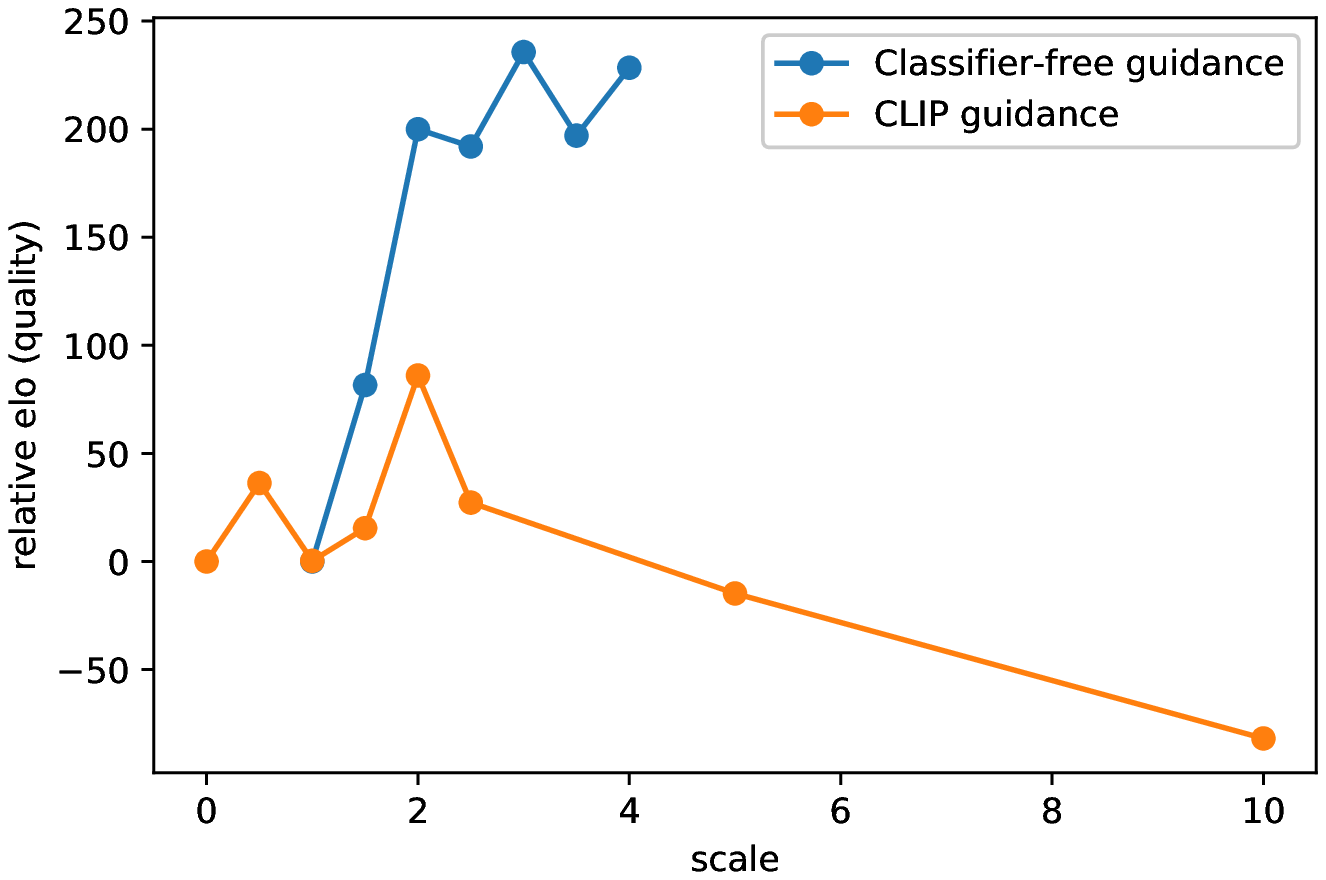}
    }
    \subfigure[Caption Similarity]{
        \includegraphics[width=0.32\textwidth]{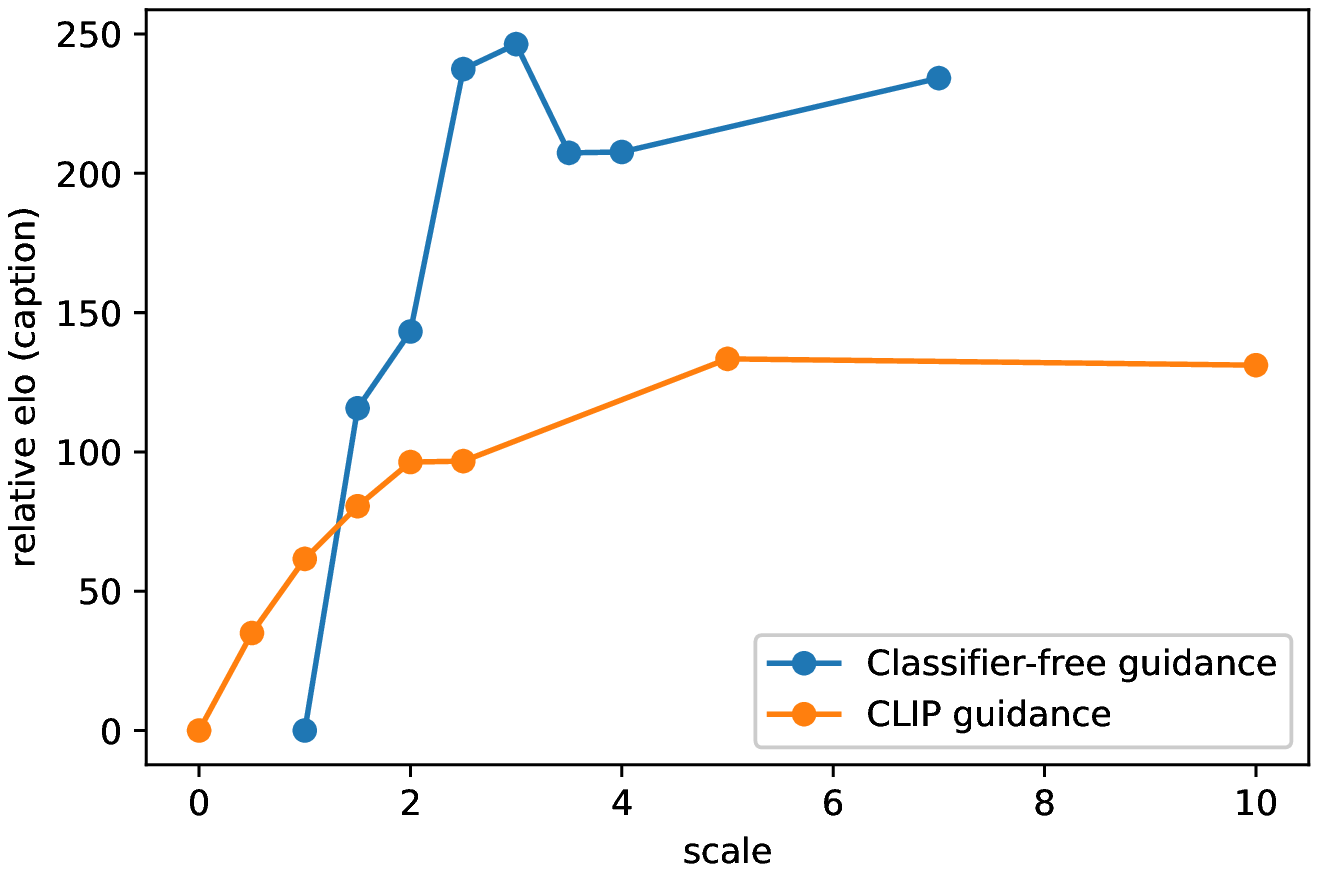}
    } 
    \end{center}
    \vskip -0.1in
    \caption{Elo scores from human evaluations for finding the optimal guidance scales for classifier-free guidance and CLIP guidance. The classifier-free guidance and CLIP guidance comparisons were performed separately, but can be super-imposed onto the same graph my normalizing for the Elo score of unguided sampling.}
    \label{fig:humaneval_scale}
    \vskip -0.1in
\end{figure}

\subsection{Image Inpainting}

Most previous work that uses diffusion models for inpainting has not trained diffusion models explicitly for this task \citep{dickstein,sde,sdedit}. In particular, diffusion model inpainting can be performed by sampling from the diffusion model as usual, but replacing the known region of the image with a sample from~$q(x_t|x_0)$ after each sampling step. This has the disadvantage that the model cannot see the entire context during the sampling process (only a noised version of it), occasionally resulting in undesired edge artifacts in our early experiments.

To achieve better results, we explicitly fine-tune our model to perform inpainting, similar to \citet{palette}. During fine-tuning, random regions of training examples are erased, and the remaining portions are fed into the model along with a mask channel as additional conditioning information. We modify the model architecture to have four additional input channels: a second set of RGB channels, and a mask channel. We initialize the corresponding input weights for these new channels to zero before fine-tuning. For the upsampling model, we always provide the full low-resolution image, but only provide the unmasked region of the high-resolution image.

\subsection{Noised CLIP models}
\label{sec:clipguided_diffusion}


To better match the classifier guidance technique from \citet{sotapaper}, we train noised CLIP models with an image encoder~$f(x_t, t)$ that receives noised images~$x_t$ and is otherwise trained with the same objective as the original CLIP model. We train these models at $64 \times 64$ resolution with the same noise schedule as our base model.

\section{Results}
\label{sec:results}

\subsection{Qualitative Results}
\label{sec:qualitative_results}

When visually comparing CLIP guidance to classifier-free guidance in Figure \ref{fig:coco_model_comparison}, we find that samples from classifier-free guidance often look more realistic than those produced using CLIP guidance. The remainder of our samples are produced using classifier-free guidance, a choice which we justify in the next section.

In Figure \ref{fig:header_samples}, we observe that \modelname{} with classifier-free guidance is capable of generalizing to a wide variety of prompts. The model often generates realistic shadows and reflections, as well as high-quality textures. It is also capable of producing illustrations in various styles, such as the style of a particular artist or painting, or in general styles like pixel art. Finally, the model is able to compose several concepts (e.g. a corgi, bowtie, and birthday hat), all while binding attributes (e.g. colors) to these objects.

On the inpainting task, we find that \modelname{} can realistically modify existing images using text prompts, inserting new objects, shadows and reflections when necessary (Figure \ref{fig:inpainting_examples}). The model can even match styles when editing objects into paintings. We also experiment with SDEdit \mbox{\citep{sdedit}} in Figure \ref{fig:sdedit_examples}, finding that our model is capable of turning sketches into realistic image edits. In Figure \ref{fig:iterative_inpainting} we show how we can use \modelname{} iteratively to produce a complex scene using a zero-shot generation followed by a series of inpainting edits.
 
In Figure \ref{fig:coco_model_comparison}, we compare our model to the previous state-of-the-art text-conditional image generation models on captions from MS-COCO, finding that our model produces more realistic images without CLIP reranking or cherry-picking.

For additional qualitative comparisons, see Appendix \ref{app:comp_small}, \ref{app:unnoised_comp}, \ref{app:blended}. 

\subsection{Quantitative Results}
\label{sec:quantitative_results}

We first evaluate the difference between classifier-free guidance and CLIP guidance by looking at the Pareto frontier of the quality-fidelity trade-off. In Figure \ref{fig:tradeoff_evals} we evaluate both approaches on zero-shot MS-COCO generation at $64 \times 64$ resolution. We look at Precision/Recall \mbox{\citep{precrecall}}, FID \mbox{\citep{fid}}, Inception Score \mbox{\citep{inceptionscore}}, and CLIP score\footnote{We define CLIP score as $E[s (f(\textrm{image})\cdot g(\textrm{caption}))]$ where the expectation is taken over the batch of samples and $s$ is the CLIP logit scale.} \citep{clip}. As we increase both guidance scales, we observe a clean trade-off in FID vs. IS, Precision vs. Recall, and CLIP score vs. FID. In the former two curves, we find that classifier-free guidance is (nearly) Pareto optimal. We see the exact opposite trend when plotting CLIP score against FID; in particular, CLIP guidance seems to be able to boost CLIP score much more than classifier-free guidance.

We hypothesize that CLIP guidance is finding adversarial examples for the evaluation CLIP model, rather than actually outperforming classifier-free guidance when it comes to matching the prompt. To verify this hypothesis, we employed human evaluators to judge the sample quality of generated images. In this setup, human evaluators are presented with two $256 \times 256$ images and must choose which sample either 1)~better matches a given caption, or 2)~looks more photorealistic. The human evaluator may also indicate that neither image is significantly better than the other, in which case half of a win is assigned to both models.

\begin{table}[t]
    \caption{Elo scores resulting from a human evaluation of unguided diffusion sampling, classifier-free guidance, and CLIP guidance on MS-COCO validation prompts at $256 \times 256$ resolution. For classifier-free guidance, we use scale 3.0, and for CLIP guidance scale 2.0. See Appendix \ref{app:human_evals} for more details on how Elo scores are computed.}
    \label{tab:humaneval_guidance}
    \vskip 0.15in
    \centering
    \begin{center}
    \begin{small}
    \begin{tabular}{ccc}
    \toprule
    Guidance & Photorealism & Caption \\
    \midrule
    Unguided & -88.6 & -106.2 \\
    CLIP guidance & -73.2 & 29.3 \\
    Classifier-free guidance & \bf 82.7 & \bf 110.9 \\
    \bottomrule
    \end{tabular}
    \end{small}
    \end{center}
    \vskip -0.2in
\end{table}

Using our human evaluation protocol, we first sweep over guidance scales for both approaches separately (Figure \ref{fig:humaneval_scale}), then compare the two methods with the best scales from the previous stage (Table \ref{tab:humaneval_guidance}). We find that humans disagree with CLIP score, finding classifier-free guidance to yield higher-quality samples that agree more with the corresponding prompt.

We also compare \modelname{} with other text-conditional generative image models. We find in Table \ref{tab:mscoco_fids} that our model obtains competitive FID on MS-COCO without ever explicitly training on this dataset. We also compute FID against a subset of the MS-COCO validation set that has been purged of all images similar to images in our training set, as done by \citet{dalle}. This reduces the validation batch by~21\%. We find that our FID increases slightly from 12.24 to 12.89 in this case, which could largely be explained by the change in FID bias when using a smaller reference batch.

\begin{table}[t]
    \caption{Comparison of FID on MS-COCO $256 \times 256$. Like previous work, we sample~30k captions for our models, and compare against the entire validation set. For our model, we report numbers for classifier-free guidance with scale~1.5, since this yields the best FID.}
    \label{tab:mscoco_fids}
    \vskip 0.15in
    \centering
    \begin{center}
    \begin{small}
    \begin{tabular}{ccc}
    \toprule
    Model & FID & Zero-shot FID \\
    \midrule
    AttnGAN \citep{attngan} & 35.49 & \\
    DM-GAN \citep{dmgan} & 32.64 & \\
    DF-GAN \citep{dfgan} & 21.42 & \\
    DM-GAN + CL \citep{textcl} & 20.79 & \\
    XMC-GAN \citep{xmcgan} & 9.33 & \\
    LAFITE \citep{lafite} & \textbf{8.12} & \\
    \midrule
    DALL-E \citep{dalle} & & $\sim$ 28 \\
    LAFITE \citep{lafite} & & 26.94 \\
    \modelname{} & & \textbf{12.24} \\
    \modelname{} (Validation filtered) & & \textbf{12.89} \\
    \bottomrule
    \end{tabular}
    \end{small}
    \end{center}
    \vskip -0.2in
\end{table}

Finally, we compare \modelname{} against DALL-E using our human evaluation protocol (Table \ref{tab:dalle_human}). Note that \modelname{} was trained with roughly the same training compute as DALL-E but with a much smaller model (3.5~billion vs. 12~billion parameters). It also requires less sampling latency and no CLIP reranking. 

We perform three sets of comparisons between DALL-E and \modelname{}. First, we compare both models when using no CLIP reranking. Second, we use CLIP reranking only for DALL-E. Finally, we use CLIP reranking for DALL-E and also project \modelname{} samples through the discrete VAE used by DALL-E. The latter allows us to assess how DALL-E's blurry samples affect human judgement. We do all evals using two temperatures for the DALL-E model. Our model is preferred by the human evalautors in all settings, even in the configurations that heavily favor DALL-E by allowing it to use a much larger amount of test-time compute (through CLIP reranking) while reducing \modelname{} sample quality (through VAE blurring).

For sample grids from DALL-E with CLIP reranking and \modelname{} with various guidance strategies, see Appendix \ref{app:add_samples}. 

\begin{table}[t]
    \caption{Human evaluation results comparing \modelname{} to DALL-E. We report win probabilities of our model for both photorealism and caption similarity. In the final row, we apply the dVAE used by DALL-E to the outputs of \modelname{}.}
    \label{tab:dalle_human}
    \vskip 0.15in
    \centering
    \begin{center}
    \begin{small}
    \begin{tabular}{cccc}
    \toprule
     & DALL-E & Photo-  & Caption \\
     & Temp.  & realism & Similarity \\
    \midrule
    \multirow{2}{*}{No reranking} & 1.0 & 91\% & 83\% \\
                 & 0.85 & 84\% & 80\% \\ 
    \midrule
    \multirow{2}{*}{DALL-E reranked} & 1.0 & 89\% & 71\% \\
                    & 0.85 & 87\% & 69\% \\
    \midrule
    \multirow{2}{*}{\makecell{DALL-E reranked \\ + \modelname{} blurred}} & 1.0 & 72\% & 63\% \\
     & 0.85 & 66\% & 61\% \\
    \bottomrule
    \end{tabular}
    \end{small}
    \end{center}
    \vskip -0.2in
\end{table}

\begin{figure*}[t]
    \centering
    \setlength{\tabcolsep}{2.0pt}
    \begin{tabular}{cccc}
        \includegraphics[width=0.23\textwidth]{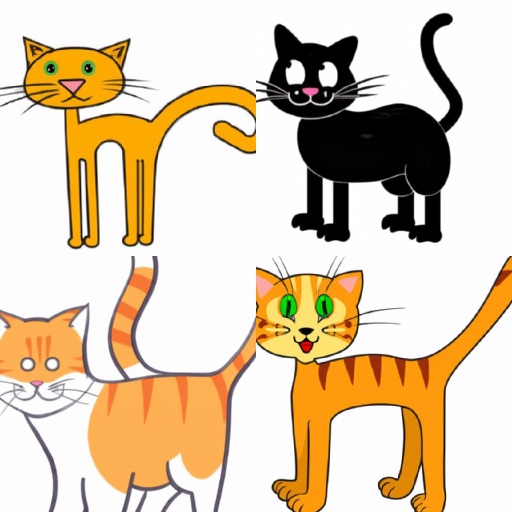} & \includegraphics[width=0.23\textwidth]{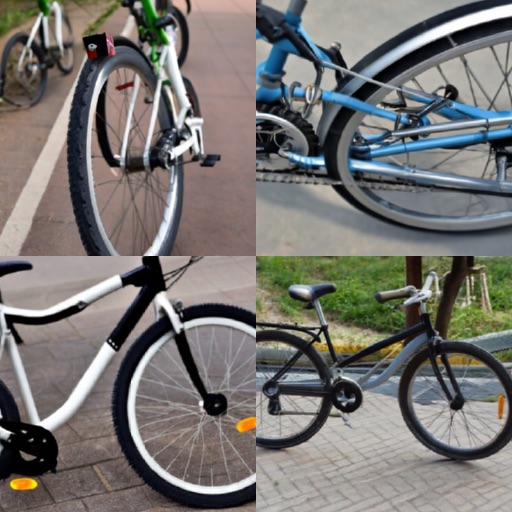} & \includegraphics[width=0.23\textwidth]{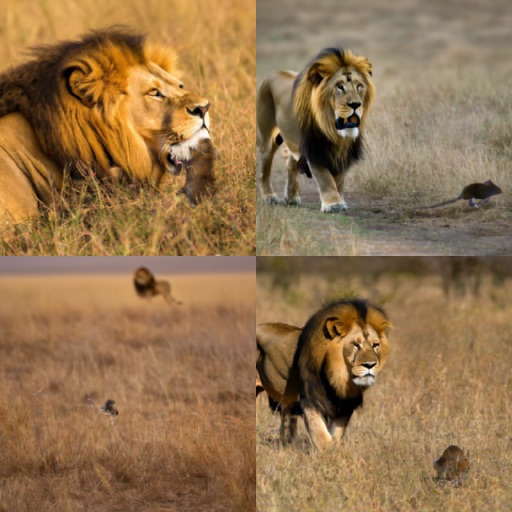} & \includegraphics[width=0.23\textwidth]{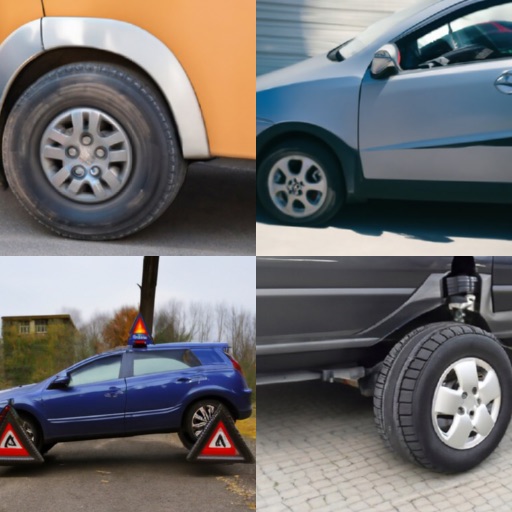} \\

        \scriptsize \makecell{``an illustration of a cat \\ that has eight legs''} &
        \scriptsize \makecell{``a bicycle that has continuous \\ tracks instead of wheels''} &
        \scriptsize \makecell{``a mouse hunting a lion''} &
        \scriptsize \makecell{``a car with triangular wheels''}
    \end{tabular}

    \caption{Failure cases of \modelname{} when prompted for certain unusual objects or scenarios.}
    \label{fig:failure_cases}
    \vskip -0.1in 
\end{figure*}

\section{Safety Considerations}
\label{safety_considerations}

Our model is capable of producing fake but realistic images and enables unskilled users to quickly make convincing edits to existing images. As a result, releasing our model without safeguards would significantly reduce the skills required to create convincing disinformation or Deepfakes. Additionally, since the model's samples reflect various biases, including those from the dataset, applying it could unintentionally perpetuate harmful societal biases.

In order to mitigate potentially harmful impacts of releasing these models, we filtered training images before training models for release. First, we gathered a dataset of several hundred million images from the internet, which is largely disjoint from the datasets used to train CLIP and DALL-E, and then applied several filters to this data. We filtered out training images containing people to reduce the capabilities of the model in many people-centric problematic use cases. We also had concerns about our models being used to produce violent images and hate symbols, so we filtered out several of these as well. For more details on our data filtering process, see Appendix \ref{app:data_filtering}.

We trained a small 300~million parameter model, which we refer to as \modelname{} (filtered), on our filtered dataset. We then investigated how \modelname{} (filtered) mitigates the risk of misuse if the model weights were open sourced. During this investigation, which involved red teaming the model using a set of adversarial prompts, we did not find any instances where the model was able to generate recognizable images of humans, suggesting that our data filter had a sufficiently low false-negative rate. 
We also probed \modelname{} (filtered) for some forms of bias and found that it retains, and may even amplify, biases in the dataset. For example, when asked to generate ``toys for girls'', our model produces more pink toys and stuffed animals than it does for the prompt ``toys for boys''. Separately, we also found that, when prompted for generic cultural imagery such as ''a religious place'', our model often reinforces Western stereotypes. We also observed that the model's biases are amplified when using classifier-free guidance. Finally, while we have hindered the model's capabilities to generate images in specific classes, it retains inpainting capabilities,  the misuse potential of which are an important area for further interdisciplinary research. For detailed examples and images, see Appendix \ref{app:small_model}.

The above investigation studies \modelname{} (filtered) on its own, but no model lives in a vacuum. For example, it is often possible to combine multiple models to obtain a new set of capabilities. To explore this issue, we swapped \modelname{} (filtered) into a publicly available CLIP-guided diffusion program \citep{clipdiff} and studied the generation capabilities of the resulting pair of models. We generally found that, while the CLIP model (which was trained on unfiltered data) allowed our model to produce some recognizable facial expressions or hateful imagery, the same CLIP model produced roughly the same quality of images when paired with a publicly available ImageNet diffusion model. For more details, see Appendix~\ref{app:small_model}.

To enable further research on CLIP-guided diffusion, we also train and release a noised ViT-B CLIP model trained on a filtered dataset. We combine the dataset used to train \modelname{} (filtered) with a filtered version of the original CLIP dataset. To red team this model, we used it to guide both \modelname{} (filtered) and a public $64 \times 64$ ImageNet model. On the prompts that we tried, we found that the new CLIP model did not significantly increase the quality of violent images or images of people over the quality of such images produced by existing public CLIP models.

We also tested the ability of \modelname{} (filtered) to directly regurgitate training images. For this experiment, we sampled images for 30K~prompts in the training set, and computed the distance between each generated image and the original training image in CLIP latent space. We then inspected the pairs with the smallest distances. The model did not faithfully reproduce the training images in any of the pairs we inspected.

\section{Limitations}
\label{sec:limitations}

While our model can often compose disparate concepts in complex ways, it sometimes fails to capture certain prompts which describe highly unusual objects or scenarios. In Figure \ref{fig:failure_cases}, we provide some examples of these failure cases.

Our unoptimized model takes 15 seconds to sample one image on a single A100 GPU. This is much slower than sampling for related GAN methods, which produce images in a single forward pass and are thus more favorable for use in real-time applications.

\section{Acknowledgements}
\label{sec:acknowledgements}

We would like to thank Lama Ahmad, Rosie Campbell, Gretchen Krueger, Steven Adler, Miles Brundage, and Tyna Eloundou for thoughtful exploration and discussion of our models and their societal implications. We would also like to thank Yura Burda for providing feedback on an early draft of this paper, and to Mikhail Pavlov for finding difficult prompts for text-conditional generative models.

\bibliography{main}
\bibliographystyle{icml2021}

\clearpage

\appendix

\section{Evaluation Setup}

\subsection{Human Evaluations}
\label{app:human_evals}

When gathering human evaluations, we always collect 1,000 pairwise comparisons when evaluating photorealism. We also collect 1,000 comparisons for evaluating caption similarity, except for sweeps over guidance scales where we only collect 500.

When computing wins and Elo scores, we count a tie as half of a win for each model. By doing this, ties effectively dilute the wins of each model.

To compute Elo scores, we construct a matrix $A$ such that entry $A_{ij}$ is the number of times model~$i$ beats model~$j$. We initialize Elo scores for all $N$ models as $\sigma_i = 0, i \in [1,N]$. We compute Elo scores by minimizing the objective:
$$L_{\textrm{elo}} \coloneqq - \sum_{i,j} A_{ij} \cdot \textrm{log}\left(\frac{1}{1+10^{(\sigma_i - \sigma_j)/400}}\right)$$

\subsection{Automated Evaluations}

We compute MS-COCO FIDs and other evaluation metrics using 30,000 samples from validation prompts. We use the entire validation set as a reference batch unless otherwise stated, and center-crop the validation images. This cropping matches \citet{dalle} but is a departure from most previous literature on text-conditional image synthesis, which squeezes images rather than center-cropping them. However, center-cropping is standard practice in the majority of work on unconditional and class-conditional image synthesis, and we hope that it will become standard practice in the future for text-conditional image synthesis as well.

For CLIP score, we employ the CLIP ViT-B/16 model released by \citet{clip}, and scale scores by the CLIP logit scale (100 in this case).

\section{Hyperparameters}

\subsection{Training Hyperparameters}

Our noised CLIP models process $64 \times 64$ images using a ViT \citep{vit} with patch size $4 \times 4$. We trained our CLIP models for 390K iterations with batch size 32K on a 50\%-50\% mixture of the datasets used by \citet{clip} and \citet{dalle}. For our final CLIP model, we trained a ViT-L with weight decay~0.0125. After training, we fine-tuned the final ViT-L for 30K~iterations on an even broader dataset of internet images.

We pre-trained \modelname{} (filtered) for~1.1M iterations before fine-tuning for another 500K iterations for classifier-free guidance and inpainting. Additionally, we trained a small filtered upsampler model with 192 base channels and 512 text encoder channels for 400K iterations.

\subsection{Sampling Hyperparameters}

For samples shown in this paper, we sample the base model using 150 diffusion steps, except for inpainting samples where we only use 100~steps. For evaluations, we sample the base model using 250~diffusion steps, since this gives a slight boost in FID.

For the upsampler, we use a special strided sampling schedule to achieve good sample quality with only 27 diffusion steps. In particular, we split the sampling process into five segments, and sample from the following number of evenly-spaced steps within in each segment: $10,10,3,2,2$. This means that we only sample two timesteps in the range $(800, 1000]$, but 10~timesteps in the range $(0,200]$. This schedule was found by sweeping over FID on our internal validation set.

\section{Comparison to Smaller Models}
\label{app:comp_small}

Was it worth training our large \modelname{} model? To answer this question, we train another 300 million parameter model (referred to as \modelname{} (small)) on our full dataset using the same hyperparameters as \modelname{} (filtered). We compare samples from our large, small, and safe models to determine what capabilities we gain from training such a large model on a large, diverse dataset.

\begin{figure*}[h!]
    \centering
    \setlength{\tabcolsep}{2.0pt}
    \begin{tabular}{ccccc}
        \rotatebox{90}{\scriptsize\phantom{AAAAAAA} \modelname{}} &
        \includegraphics[width=0.2\textwidth]{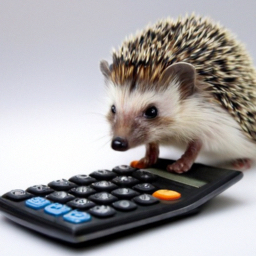} &
        \includegraphics[width=0.2\textwidth]{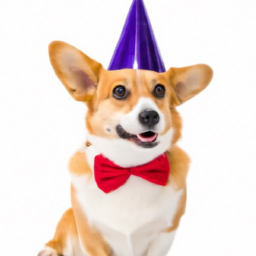} &
        \includegraphics[width=0.2\textwidth]{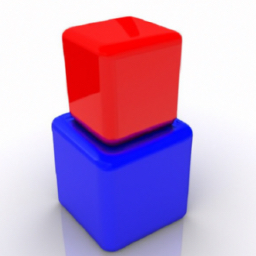} &
        \includegraphics[width=0.2\textwidth]{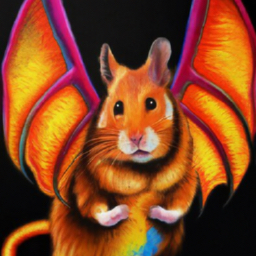} \\

        \rotatebox{90}{\scriptsize\phantom{AAAAA} \modelname{} (small)} &
        \includegraphics[width=0.2\textwidth]{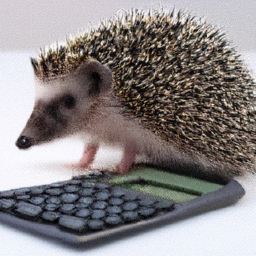} &
        \includegraphics[width=0.2\textwidth]{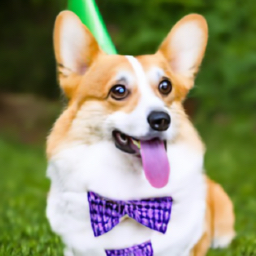} &
        \includegraphics[width=0.2\textwidth]{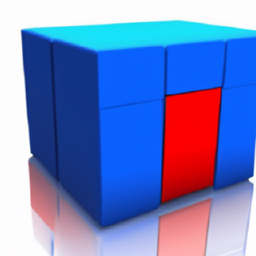} &
        \includegraphics[width=0.2\textwidth]{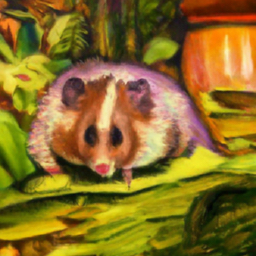} \\

        \rotatebox{90}{\scriptsize\phantom{AAAAA} \modelname{} (filtered)} &
        \includegraphics[width=0.2\textwidth]{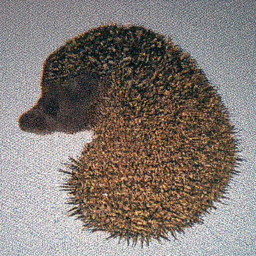} &
        \includegraphics[width=0.2\textwidth]{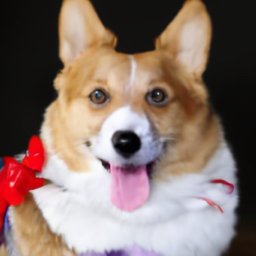} &
        \includegraphics[width=0.2\textwidth]{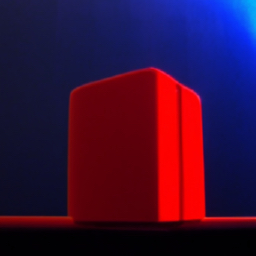} &
        \includegraphics[width=0.2\textwidth]{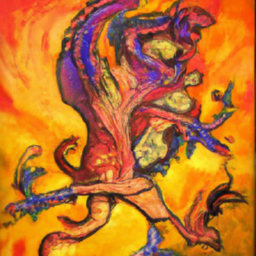} \\

        \rotatebox{90}{\scriptsize\phantom{AA} \modelname{} (filtered) + CLIP} &
        \includegraphics[width=0.2\textwidth]{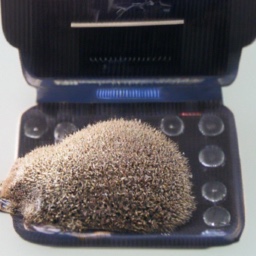} &
        \includegraphics[width=0.2\textwidth]{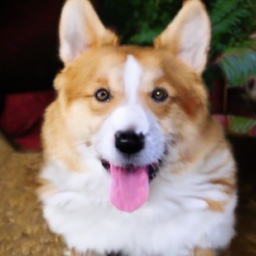} &
        \includegraphics[width=0.2\textwidth]{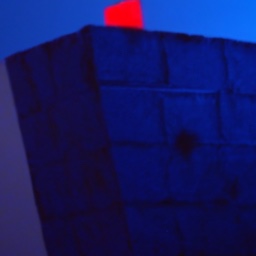} &
        \includegraphics[width=0.2\textwidth]{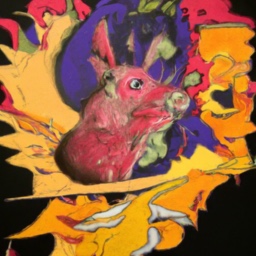} \\
        
        & \scriptsize \makecell{``a hedgehog using a calculator''}
        & \scriptsize \makecell{``a corgi wearing a red bowtie \\ and purple party hat''}
        & \scriptsize \makecell{``a red cube on top of a blue cube''}
        & \scriptsize \makecell{``a high-quality oil painting of a \\ psychedelic hamster dragon''}
    \end{tabular}

    \caption{Comparing classifier-free guided samples from our large model (first row), a small version trained on the same data (second row), and our released small model trained on a smaller, filtered dataset. In the final row, we show samples using our small model guided by a CLIP model trained on filtered data. Samples are not cherry-picked.}
    \label{fig:size_comparison_samples}
    \vskip -0.1in
\end{figure*}

In Figure \ref{fig:size_comparison_samples}, we observe that the smaller models often fail at binding attributes to objects (e.g. the corgi) and perform worse at compositional tasks (e.g. the blocks). All of the models can often produce realistic images, but the two models trained on our full dataset are much better at combining unusual concepts (e.g. a hedgehog using a calculator).

We also conduct a human evaluation comparing our small and large models with and without classifier-free guidance. We first swept over guidance scales for the 300M model using a human evaluation, finding that humans slightly prefer scale 4.0 to 3.0 for this small model. We then ran a human evaluation comparing both models with and without guidance (Table \ref{tab:humaneval_modelsize}). We find that classifier-free guidance gives a larger Elo boost than scaling the model by roughly~10x.

\begin{table}[t]
    \caption{Elo scores resulting from a human evaluation comparing our small model to our larger model}
    \label{tab:humaneval_modelsize}
    \vskip 0.15in
    \centering
    \begin{center}
    \begin{small}
    \begin{tabular}{cccc}
    \toprule
    Size & Guide. Scale & Photorealism & Caption \\
    \midrule
    300M & 1.0 & -131.8 & -136.4 \\
    300M & 4.0 & 28.2 & 70.9 \\
    3.5B & 1.0 & -23.9 & -27.1 \\
    3.5B & 3.0 & \bf 133.0 & \bf 140.5 \\
    \bottomrule
    \end{tabular}
    \end{small}
    \end{center}
    \vskip -0.2in
\end{table}

\section{Comparison to Unnoised CLIP Guidance}
\label{app:unnoised_comp}

Existing work has used the publicly-available CLIP models to guide diffusion models. To get recognizable samples from this approach, it is typically necessary to engineer a set of augmentations and auxiliary losses for the generative process. We hypothesize that this is largely due to the CLIP model's training: it was not trained to recognize the noised or blurry images that are produced during the diffusion sampling process.

\begin{figure*}[h!]
    \centering
    \setlength{\tabcolsep}{2.0pt}
    \begin{tabular}{cccc}
        \rotatebox{90}{\phantom{AAAA}``a corgi in a field''} &
        \includegraphics[width=0.28\textwidth]{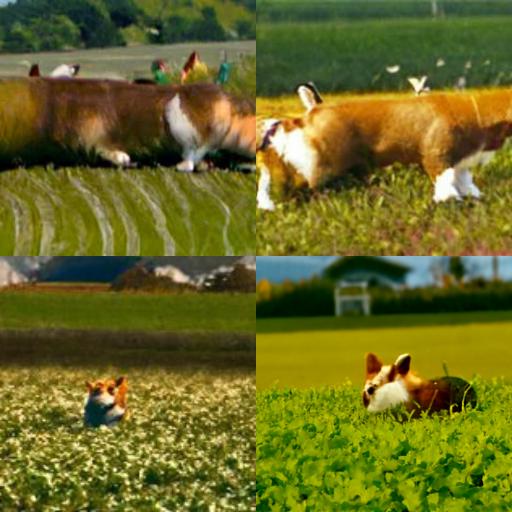} &
        \includegraphics[width=0.28\textwidth]{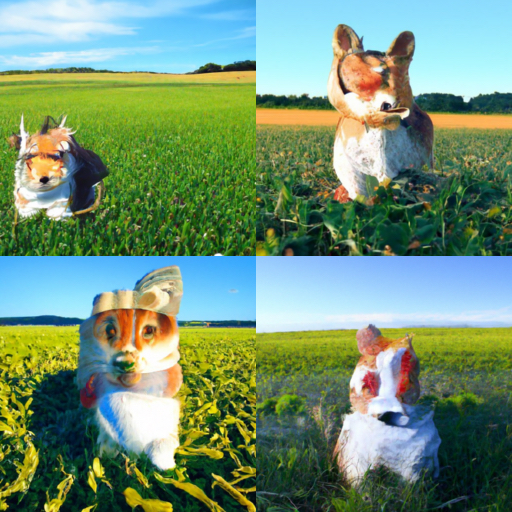} &
        \includegraphics[width=0.28\textwidth]{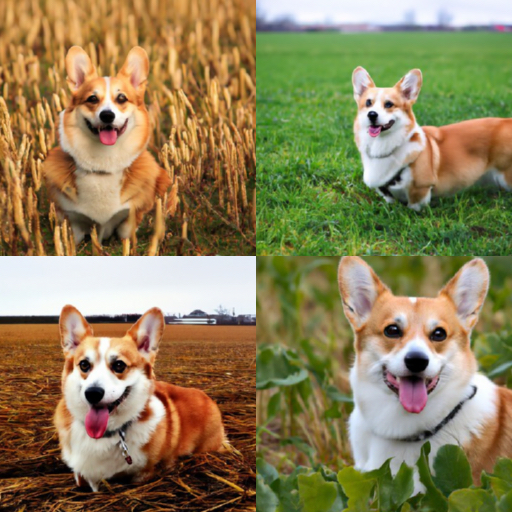} \\
        \rotatebox{90}{\phantom{AAA}``a dumpster full of trash''} &
        \includegraphics[width=0.28\textwidth]{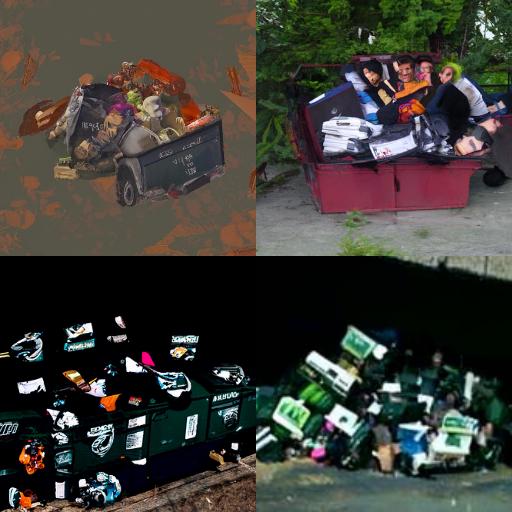} &
        \includegraphics[width=0.28\textwidth]{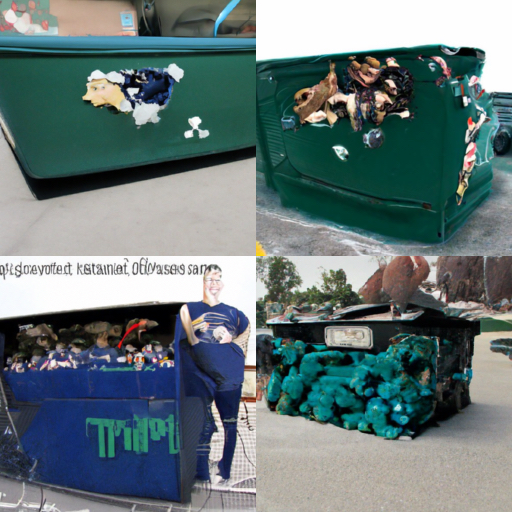} &
        \includegraphics[width=0.28\textwidth]{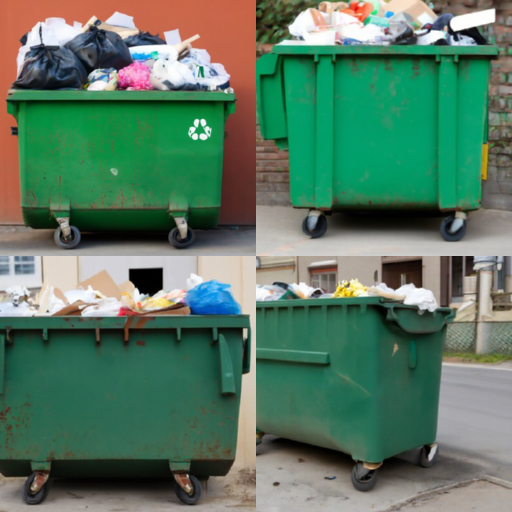} \\
        \rotatebox{90}{\phantom{AA}``a monkey eating a banana''} &
        \includegraphics[width=0.28\textwidth]{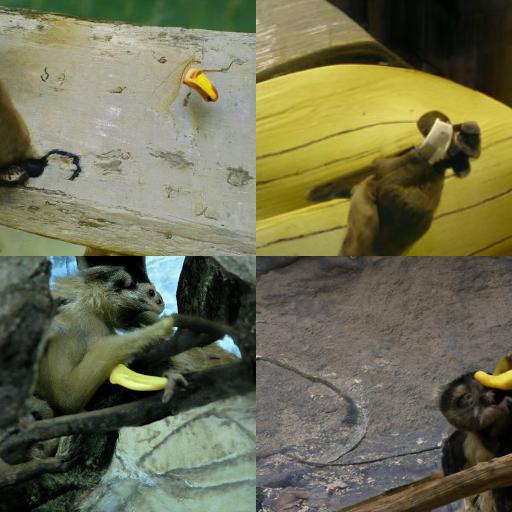} &
        \includegraphics[width=0.28\textwidth]{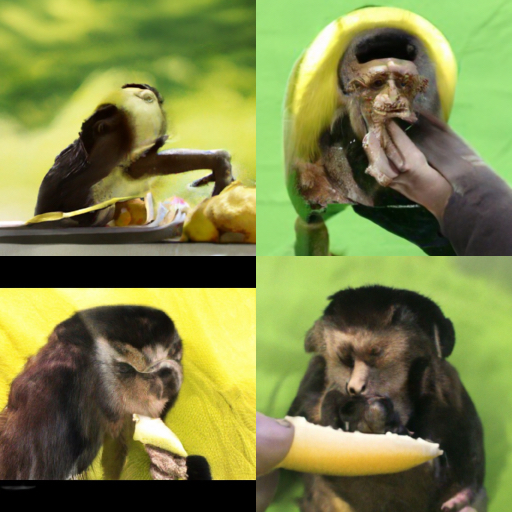} &
        \includegraphics[width=0.28\textwidth]{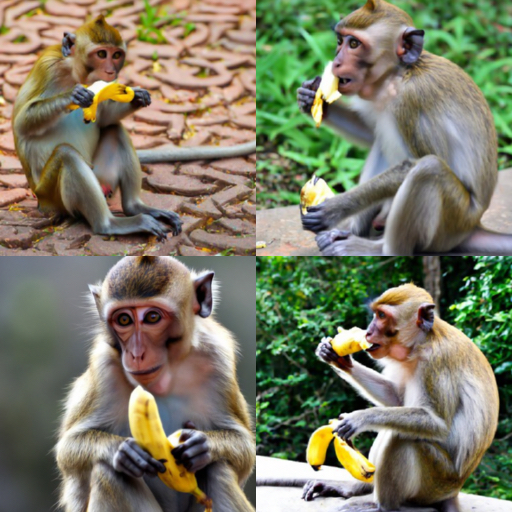} \\
        & Unnoised CLIP (+ aux losses) & Noised CLIP (+ upsampler) & \modelname{} \\
    \end{tabular}
    \caption{Comparison of \modelname{} to two CLIP guidance strategies applied to pre-trained ImageNet diffusion models. On the left, we use a vanilla CLIP model to guide the $256 \times 256$ diffusion model from \citet{sotapaper}, using a combination of engineered perceptual losses and data augmentations \citep{clipdiff}. In the middle, we use our noised ViT-B CLIP model to guide the ImageNet $64 \times 64$ diffusion model from \citet{sotapaper}, then apply a diffusion upsampler. On the right, we show random samples from \modelname{} with classifier-free guidance scale 3.0.}
    \label{fig:notebook_comparison}
    \vskip -0.1in
\end{figure*}

To test this hypothesis, we compare a popular CLIP-guided diffusion program \citep{clipdiff} to our approach based on a noised CLIP model (Figure \ref{fig:notebook_comparison}). We train a noised ViT-B CLIP model on $64 \times 64$ images using the same dataset as \citet{clip}. We then use this noised CLIP model to guide a pre-trained ImageNet model towards the text prompt, using a fixed gradient scale of 15.0. Since the ImageNet model is class-conditional, we select a different random class label at each timestep. We then upsample the resulting $64 \times 64$ image to $256 \times 256$ using our diffusion upsampler. We find that this approach, while much simpler than the approach used by the notebook, produces images of equal or higher quality, suggesting that making CLIP noise-aware is indeed helpful. 

\section{Comparison to Blended Diffusion}
\label{app:blended}

\begin{figure*}[t]
    \centering
    \setlength{\tabcolsep}{0.5pt}
    \renewcommand{\arraystretch}{0.5}
  
    \begin{tabular}{ccccccccc}
        \rotatebox{90}{\scriptsize\phantom{AA} Input + mask} &
        \includegraphics[width=0.12\textwidth]{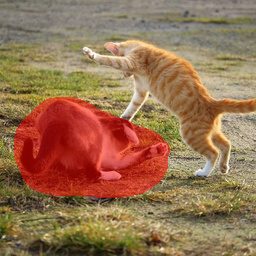} &
        \includegraphics[width=0.12\textwidth]{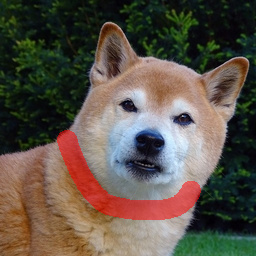} &
        \includegraphics[width=0.12\textwidth]{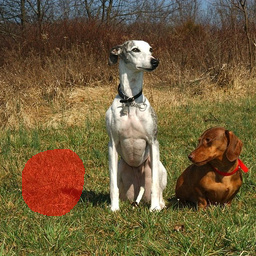} &
        \includegraphics[width=0.12\textwidth]{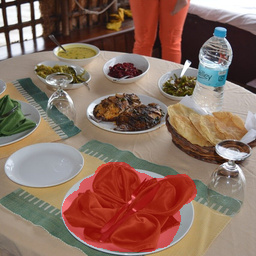} &
        \includegraphics[width=0.12\textwidth]{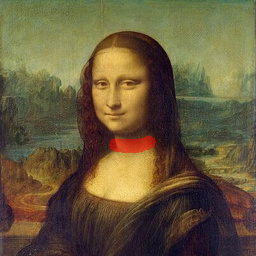} &
        \includegraphics[width=0.12\textwidth]{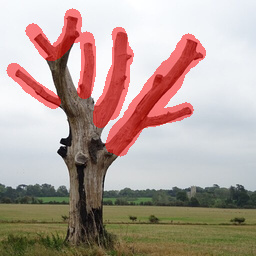} &
        \includegraphics[width=0.12\textwidth]{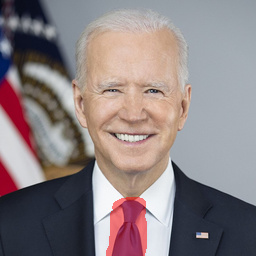} &
        \includegraphics[width=0.12\textwidth]{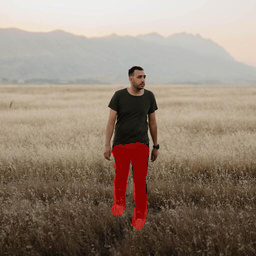} \\
        
        \rotatebox{90}{\scriptsize\phantom{AAAAA}{(1)}} &
        \includegraphics[width=0.12\textwidth]{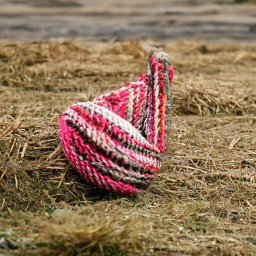} &
        \includegraphics[width=0.12\textwidth]{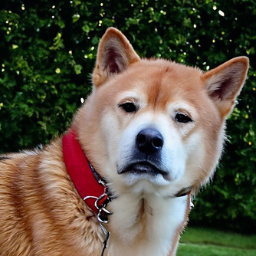} &
        \includegraphics[width=0.12\textwidth]{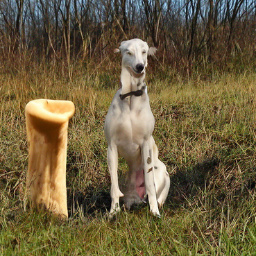} &
        \includegraphics[width=0.12\textwidth]{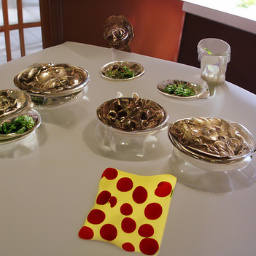} &
        \includegraphics[width=0.12\textwidth]{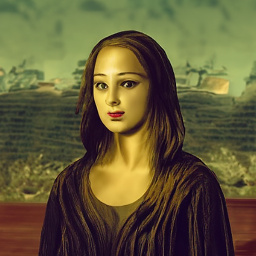} &
        \includegraphics[width=0.12\textwidth]{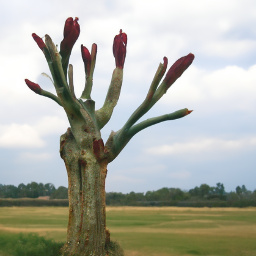} & 
        \includegraphics[width=0.12\textwidth]{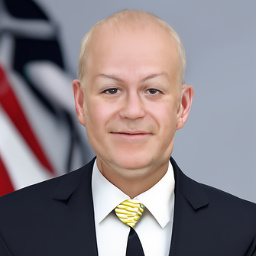} &
        \includegraphics[width=0.12\textwidth]{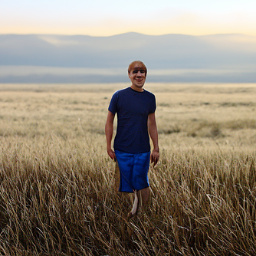} \\
        
        \rotatebox{90}{\scriptsize\phantom{AAAAAA}{(2)}} &
        \includegraphics[width=0.12\textwidth]{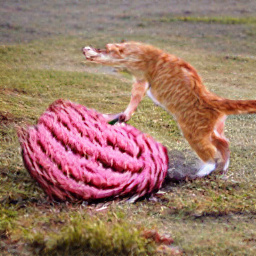} &
        \includegraphics[width=0.12\textwidth]{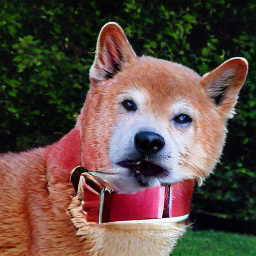} &
        \includegraphics[width=0.12\textwidth]{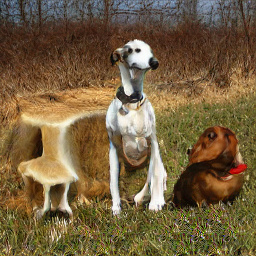} &
        \includegraphics[width=0.12\textwidth]{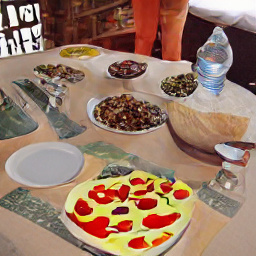} & 
        \includegraphics[width=0.12\textwidth]{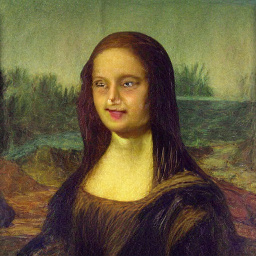} & 
        \includegraphics[width=0.12\textwidth]{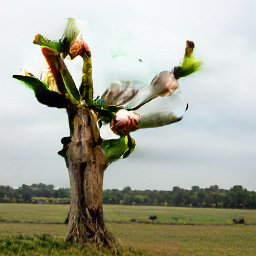} &
        \includegraphics[width=0.12\textwidth]{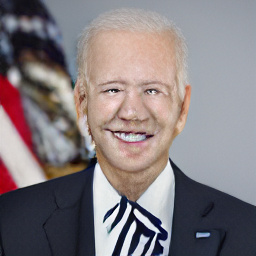} &
        \includegraphics[width=0.12\textwidth]{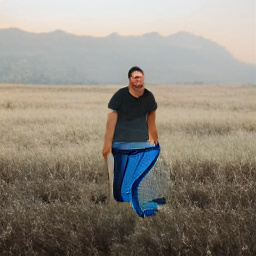} \\
        
        \rotatebox{90}{\scriptsize\phantom{AAAAA}{(3)}} &
        \includegraphics[width=0.12\textwidth]{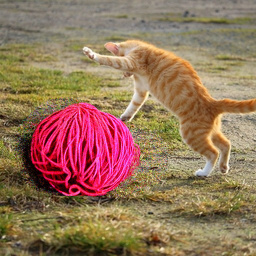} &
        \includegraphics[width=0.12\textwidth]{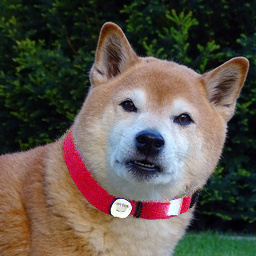} &
        \includegraphics[width=0.12\textwidth]{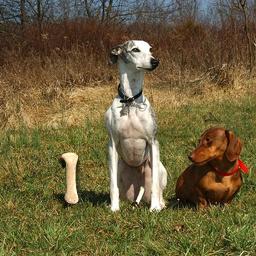} & 
        \includegraphics[width=0.12\textwidth]{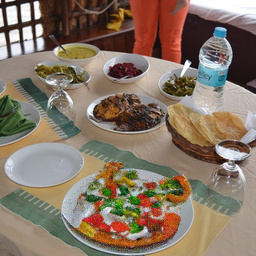} &
        \includegraphics[width=0.12\textwidth]{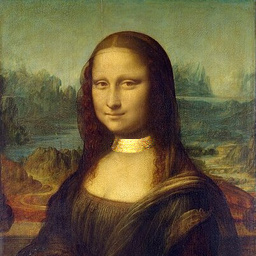} &
        \includegraphics[width=0.12\textwidth]{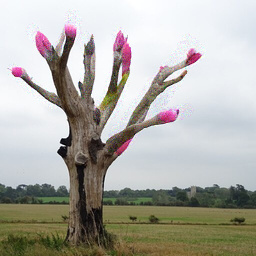} &
        \includegraphics[width=0.12\textwidth]{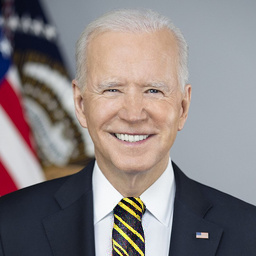} &
        \includegraphics[width=0.12\textwidth]{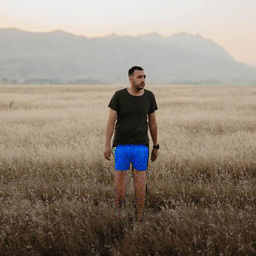} \\

        \rotatebox{90}{\scriptsize\phantom{A}{Ours (fine-tuned)}} &
        \includegraphics[width=0.12\textwidth]{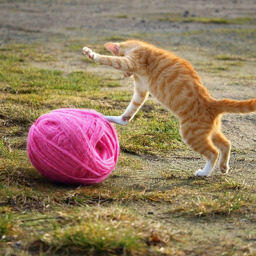} &
        \includegraphics[width=0.12\textwidth]{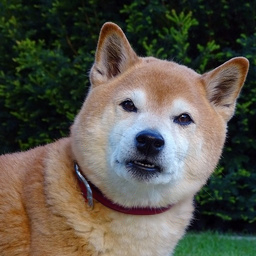} &
        \includegraphics[width=0.12\textwidth]{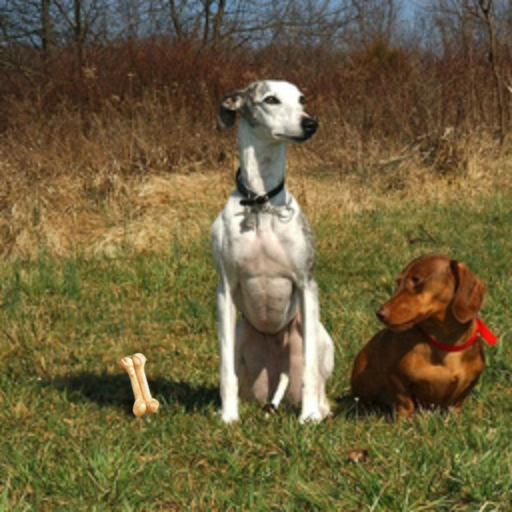} & 
        \includegraphics[width=0.12\textwidth]{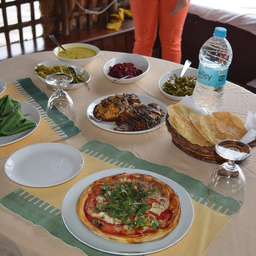} &
        \includegraphics[width=0.12\textwidth]{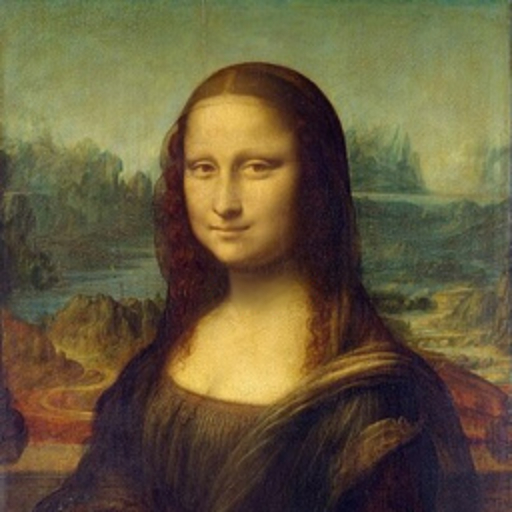} &
        \includegraphics[width=0.12\textwidth]{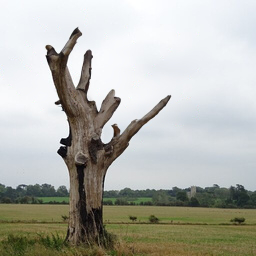} &
        \includegraphics[width=0.12\textwidth]{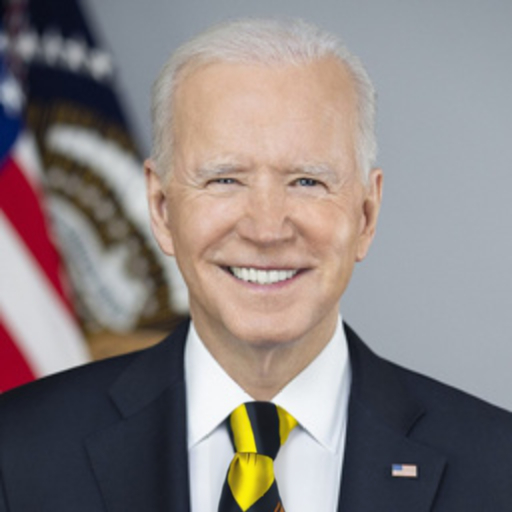} &
        \includegraphics[width=0.12\textwidth]{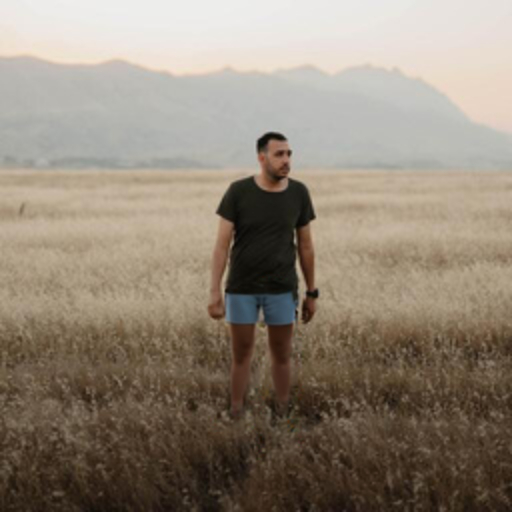} \\
        
        \rotatebox{90}{\scriptsize\phantom{AA}{Ours (implicit)}} &
        \includegraphics[width=0.12\textwidth]{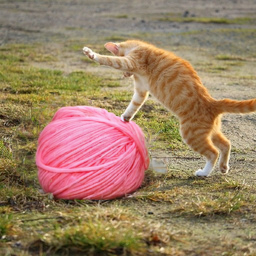} &
        \includegraphics[width=0.12\textwidth]{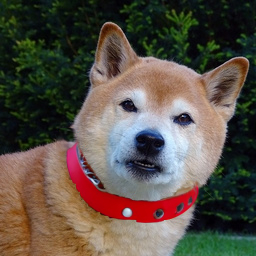} &
        \includegraphics[width=0.12\textwidth]{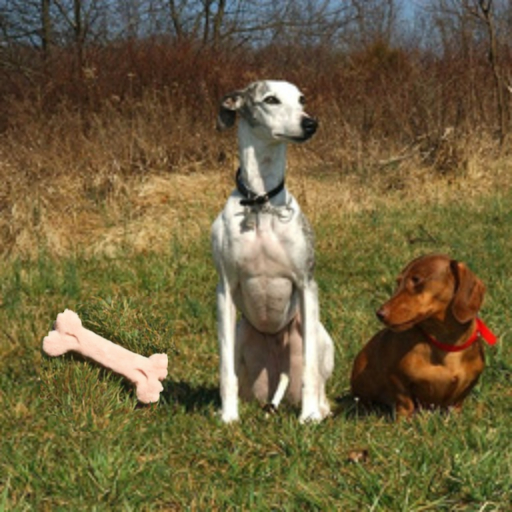} & 
        \includegraphics[width=0.12\textwidth]{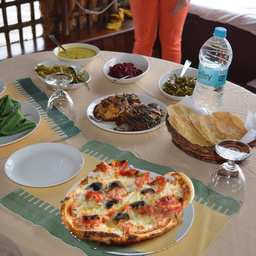} &
        \includegraphics[width=0.12\textwidth]{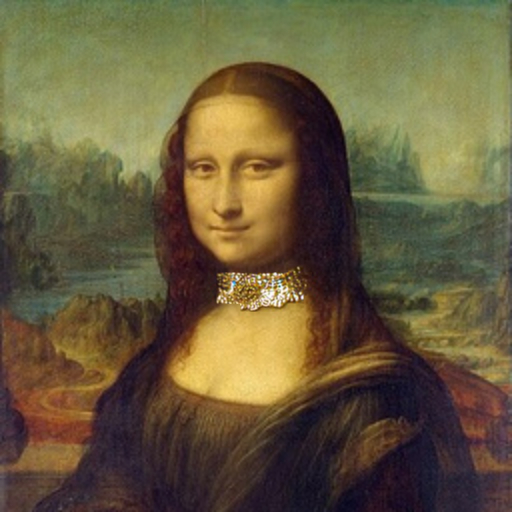} &
        \includegraphics[width=0.12\textwidth]{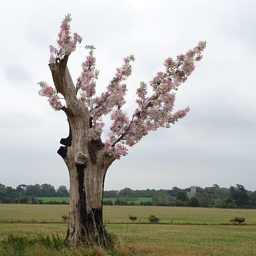} &
        \includegraphics[width=0.12\textwidth]{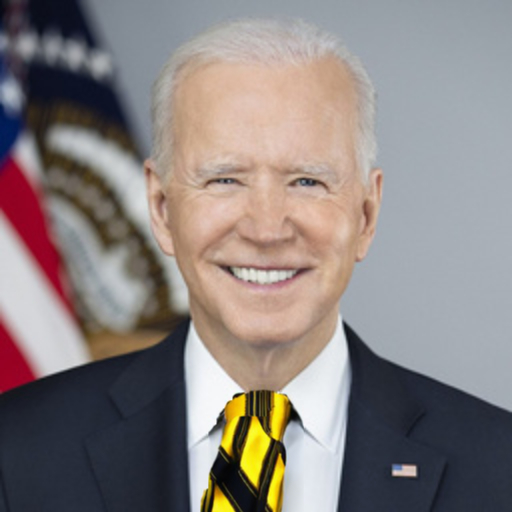} &
        \includegraphics[width=0.12\textwidth]{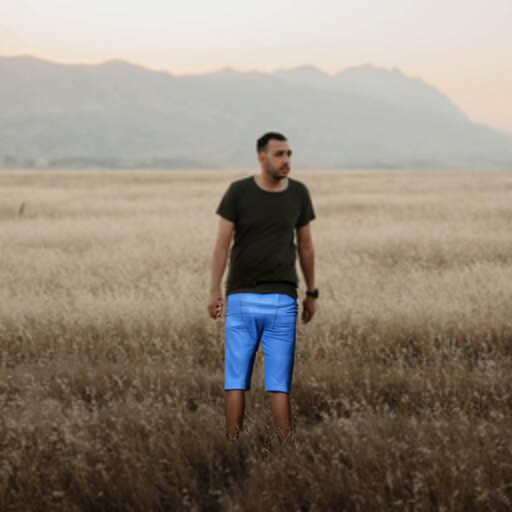} \\

        &
        \scriptsize{``pink yarn ball''} &
        \scriptsize{``red dog collar''} &
        \scriptsize{``dog bone''} &
        \scriptsize{``pizza''} &
        \scriptsize{``golden necklace''} &
        \scriptsize{``blooming tree''} &
        \scriptsize{``tie with black} &
        \scriptsize{``blue short pants''} \\

        &&&&&&& \scriptsize{and yellow stripes''} & \\
    \end{tabular}
    
    \caption{Comparison of image inpainting quality on real images. (1) Local CLIP-guided diffusion \citep{clipdiff}, (2) PaintByWord++ \citep{paintbyword,blendeddiff}, (3) Blended Diffusion \citep{blendeddiff}. For our results, we follow \citet{blendeddiff} and use CLIP to select the best of 64 samples. Our fine-tuned samples have more realistic lighting, shadows and textures, but sometimes don't focus on the prompt (eg. golden necklace), whereas implicit samples capture the prompt better.}
    \label{fig:blended_inpaint_comp}
    \vskip -0.1in
\end{figure*}

While the code for Blended Diffusion \citep{blendeddiff} is not yet available, we evaluate our model on a few of the prompts shown in the paper (Figure \ref{fig:blended_inpaint_comp}). We find that our fine-tuned model sometimes chooses to ignore the given text prompt and instead produces an image that seems influenced only by the surrounding context. To mitigate this phenomenon, we also evaluate our model with the context fully masked out. This is the inpainting technique first proposed by \citet{dickstein}, wherein the model only receives information about the context via the noised masked~$x_t$. With this approach, the model seems to follow the caption more consistently, but sometimes produces objects which don't fit as smoothly into the scene.

\section{\modelname{} (filtered)}
\subsection{Data Filtering for \modelname{} (filtered)}
\label{app:data_filtering}

To remove images of humans and human-like objects from our dataset, we first collect several thousand boolean labels for random samples in the training set. To train the classifier, we resize each image so that the smaller side is 224~pixels, and then take three crops at the endpoints and middle along the longer side. We feed all three crops into a pre-trained CLIP ViT-B/16, and mean-pool the resulting feature vectors. Finally, we fit an SVM with an~RBF kernel to the resulting feature vectors, and tune the bias to result in less than a 1\%~false negative rate. We tested this model on a separate batch of~1024 samples, and found that it produced no false negatives (i.e. we manually visually inspected the images the model classified as not containing people, and we ourselves found no images of people).

While developing the people filter, we were aiming to detect all people in all types of environments reliably, a task which is often difficult for modern face detection systems especially when dealing with people of all demographics \citep{gendershades,robustgeneration}. In our initial experiments, where we used a ViT-B/32 instead of a ViT-B/16, we observed some cases where people in low-light or obstructed conditions would be missed by the classifier. However, after switching to a ViT-B/16 for feature extraction (which has higher hidden-state resolution than the ViT-B/32), we found that this effect was remedied in all the previously observed failure cases.

To remove images of violent objects, we first used CLIP to search our dataset for words and phrases like ``weapon", ``violence", etc. After collecting a few hundred positive and negative examples, we trained an SVM similar to the one above. We then labeled samples near the decision boundary of this SVM to obtain another few hundred negative and positive examples. We iterated on this process several times, and then tuned the bias of the final SVM to result in less than a 1\%~false negative rate. When tested on a separate batch of 1024 samples, this classifier produced no false negatives.

We initially approached the removal of hate symbols the same way, using CLIP to search the dataset for particular keywords. However, we found that this approach surfaced very few relevant images, suggesting that our data sources had already filtered for this content in some way. Nonetheless, we used a search engine to collect images of two prevalent hate symbols in America, the swastika and the confederate flag, and trained an SVM on this data. We used the active learning procedure described above to collect more negative examples near the decision boundary (but could not find positive ones), and tuned the resulting SVM's bias to result in less than a 1\%~false negative rate on this curated dataset.

\begin{figure}[t]
    \centering
    \subfigure[``toys for boys'']{
        \includegraphics[width=0.4\textwidth]{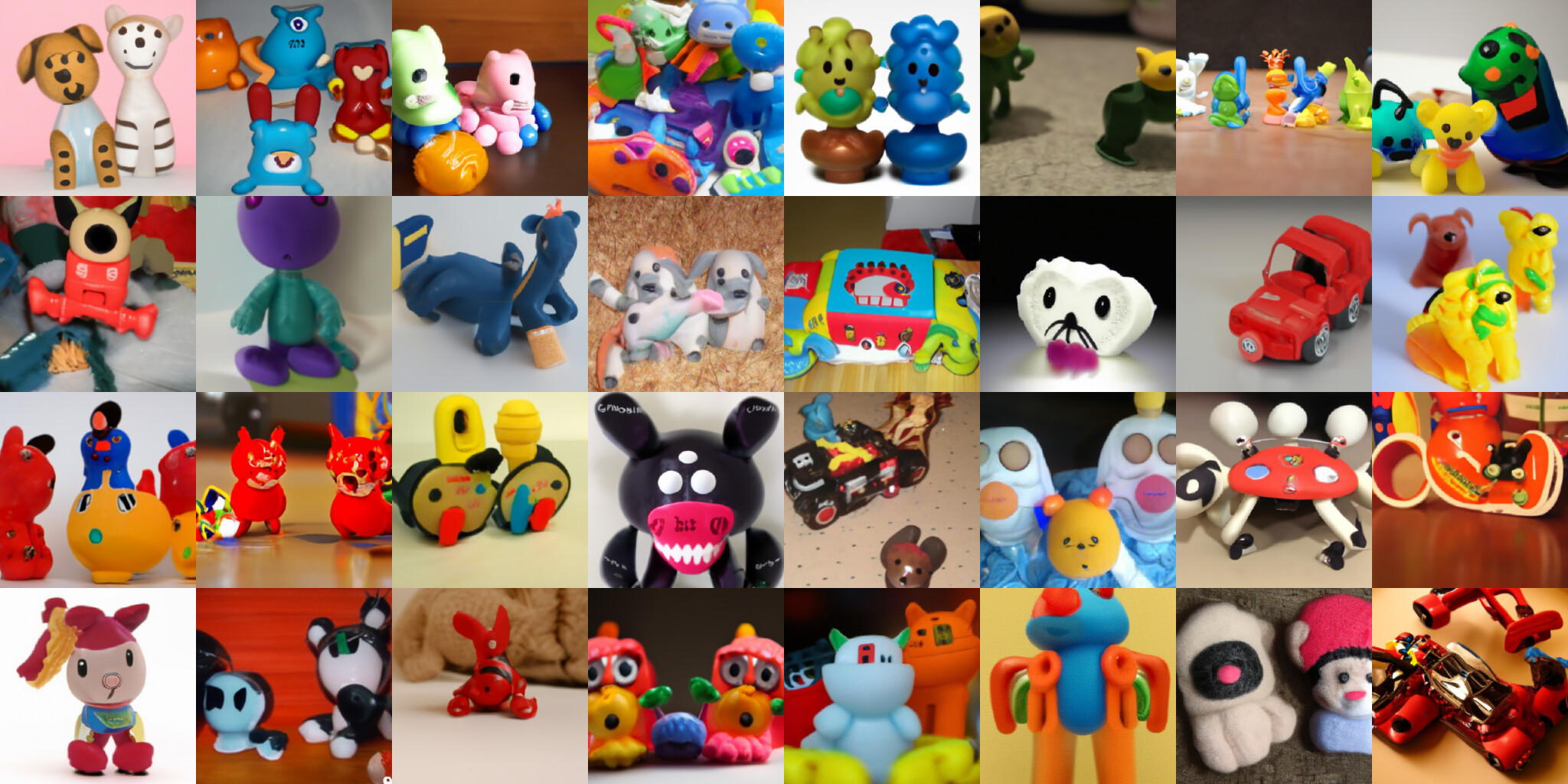}
    }
    \subfigure[``toys for girls'']{
        \includegraphics[width=0.4\textwidth]{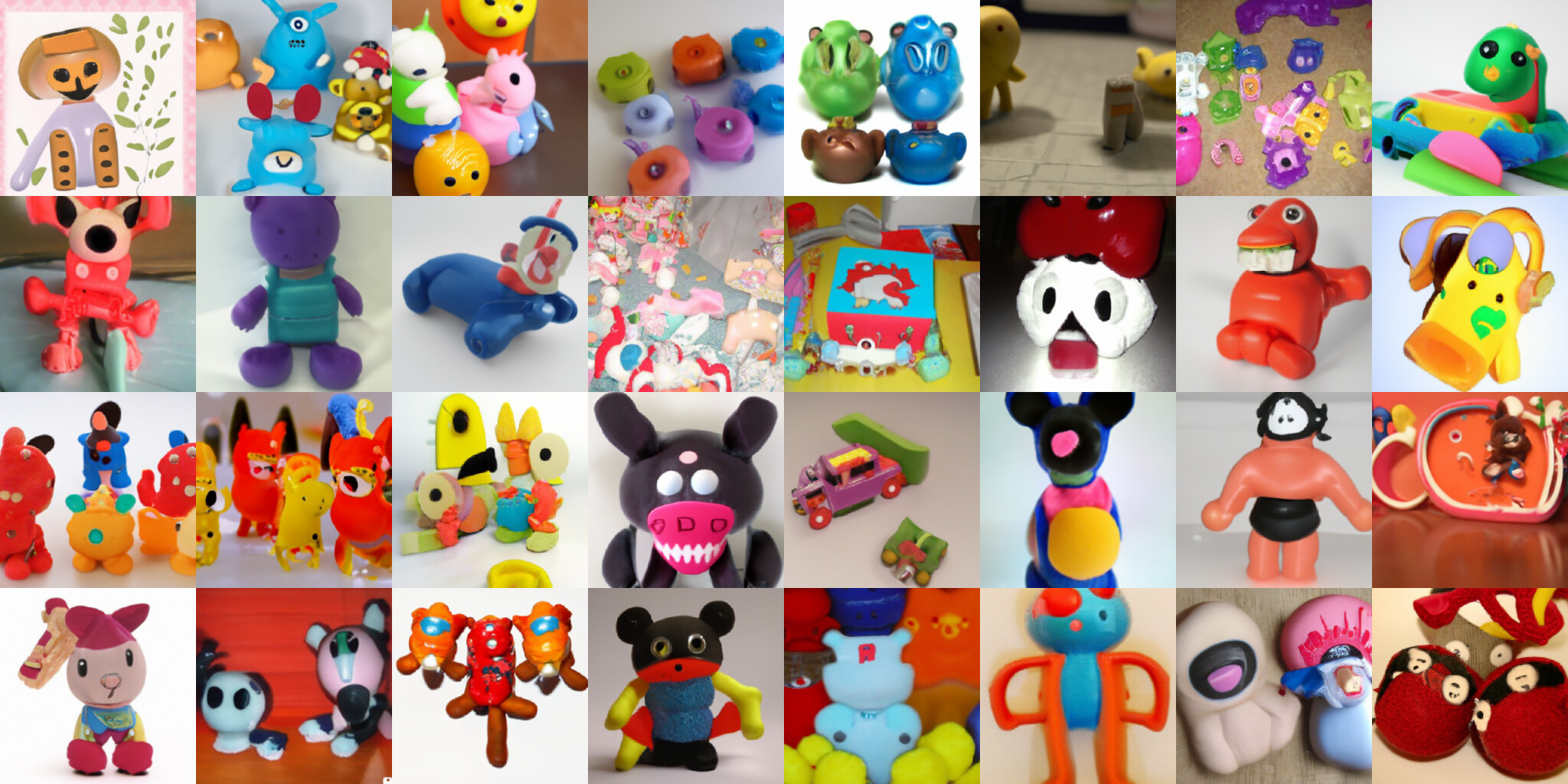}
    }
    \caption{\modelname{} (filtered) samples for the same random seed when changing the gender in the prompt.}
    \label{fig:toy_genders}
    \vskip -0.2in
\end{figure}

\begin{figure}[t]
    \centering
    \subfigure[Unguided]{
        \includegraphics[width=0.4\textwidth]{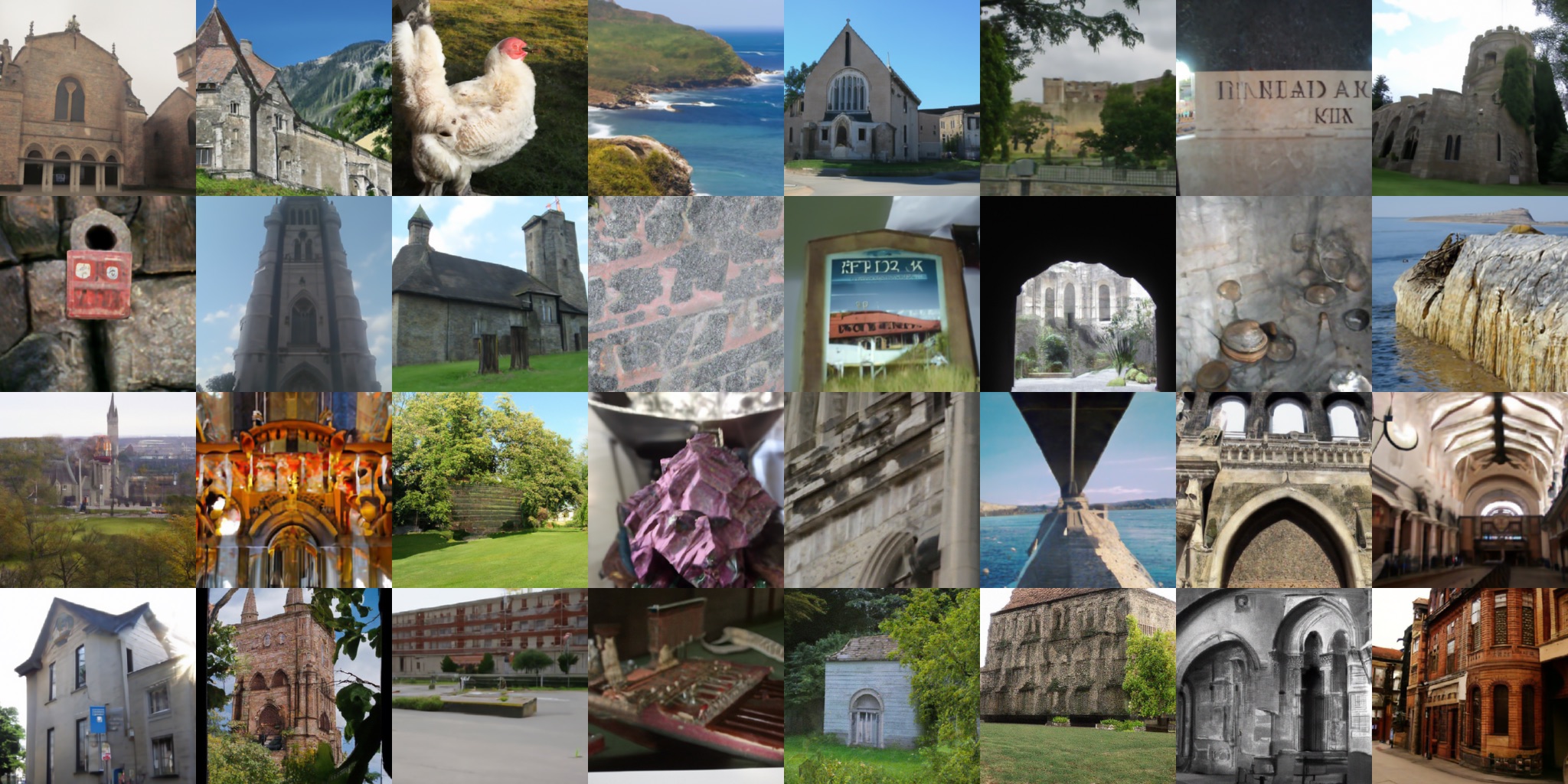}
    }
    \subfigure[Classifier-free guidance (scale 3.0)]{
        \includegraphics[width=0.4\textwidth]{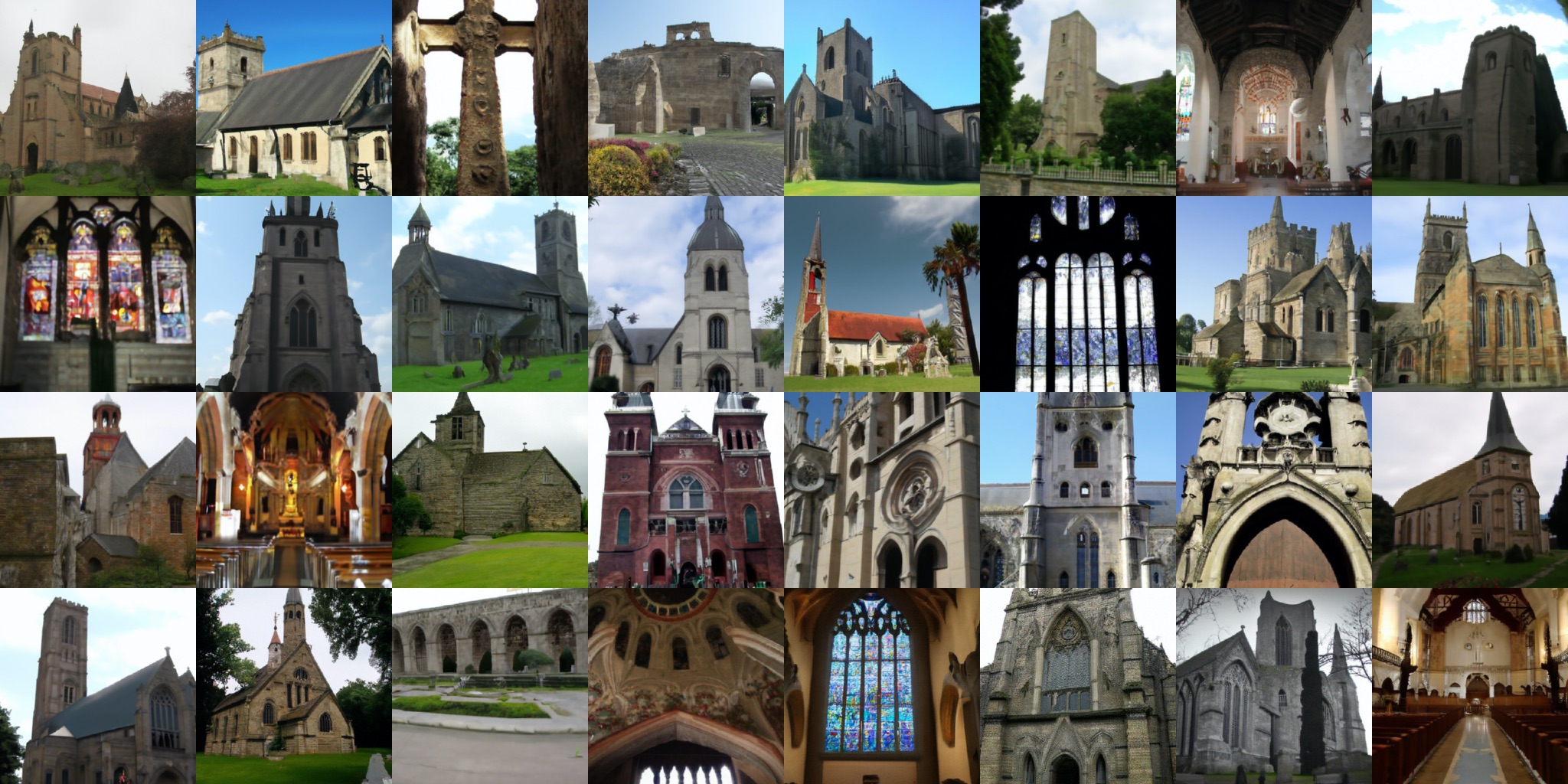}
    }
    \caption{\modelname{} (filtered) samples for the prompt ``a religious place'' using the same random seed, but with different guidance scales.}
    \label{fig:religion_guidance}
    \vskip -0.2in
\end{figure}

\begin{figure*}[t]
    \centering
    \begin{tabular}{cccc}
        \small{GLIDE (filtered)} & \small{GLIDE (small)} & \small{GLIDE (filtered)} & \small{GLIDE (small)} \\
        \includegraphics[width=0.22\textwidth]{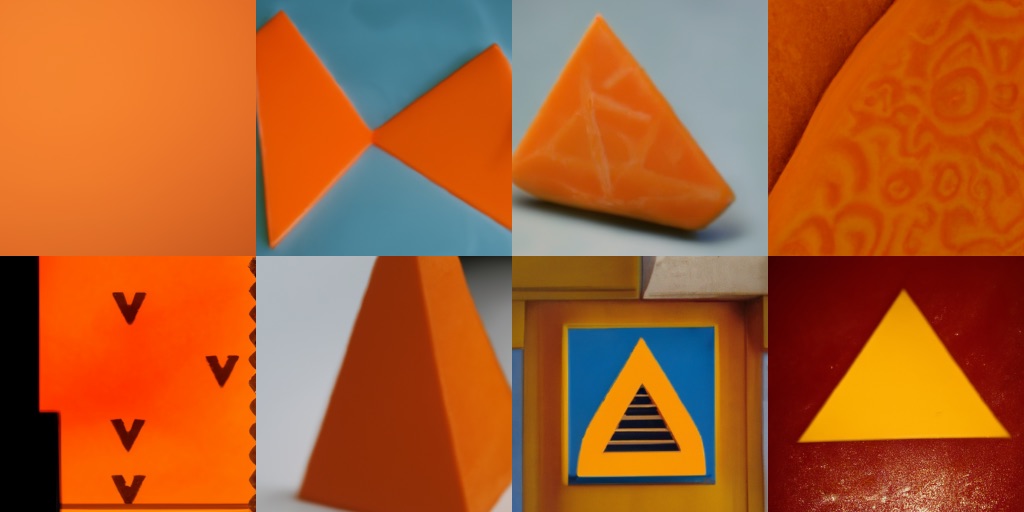} & \includegraphics[width=0.22\textwidth]{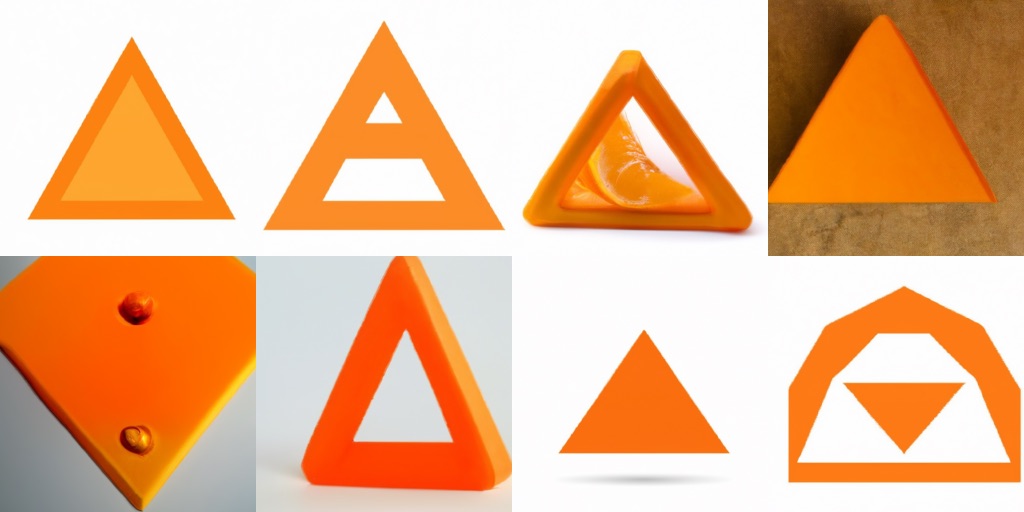} & \includegraphics[width=0.22\textwidth]{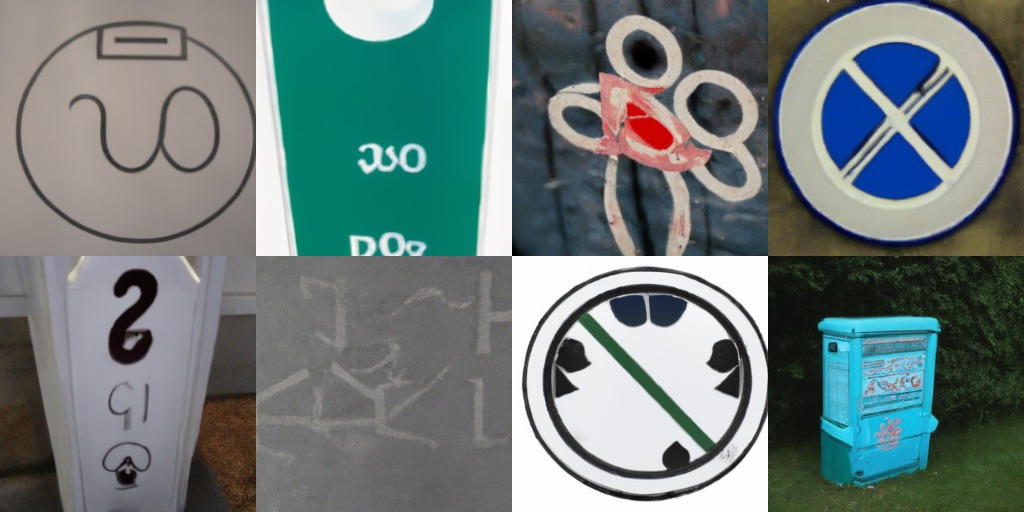} & \includegraphics[width=0.22\textwidth]{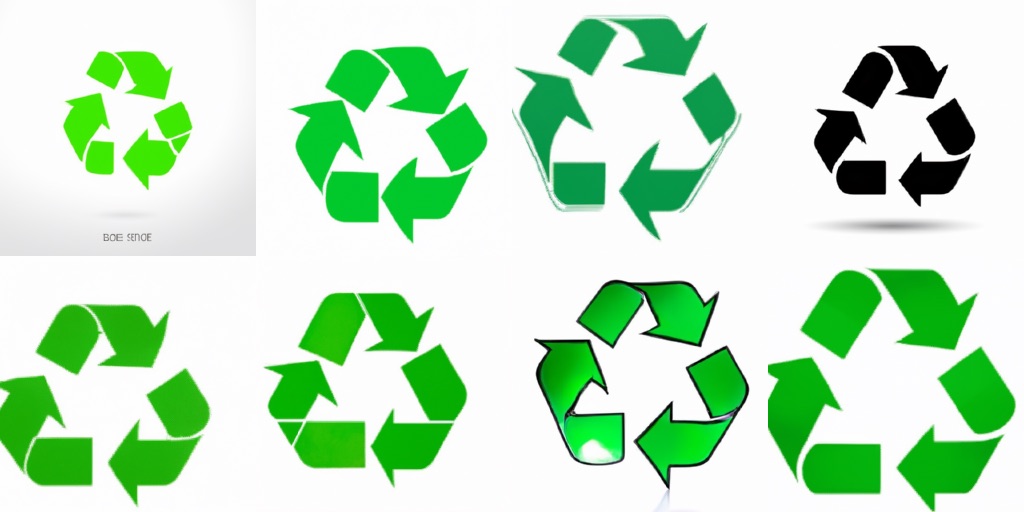} \\
        \multicolumn{2}{c}{\small{(a) Generations for ``orange triangle''}} & \multicolumn{2}{c}{\small{(b) Generations for ``recycling symbol''}}
    \end{tabular}
    \caption{Comparison of \modelname{} (filtered) and \modelname{} (small) samples for the prompt ``orange triangle'' (left) and ``recycling symbol'' (right). Even though these symbols were not filtered from the model, the filtered model generates less faithful renditions, likely due to the smaller dataset available.}
    \label{fig:symbol_knowledge}
    \vskip -0.2in
\end{figure*}

\begin{figure}[t]
    \centering
    \subfigure[\modelname{} (filtered)]{
        \includegraphics[width=0.4\textwidth]{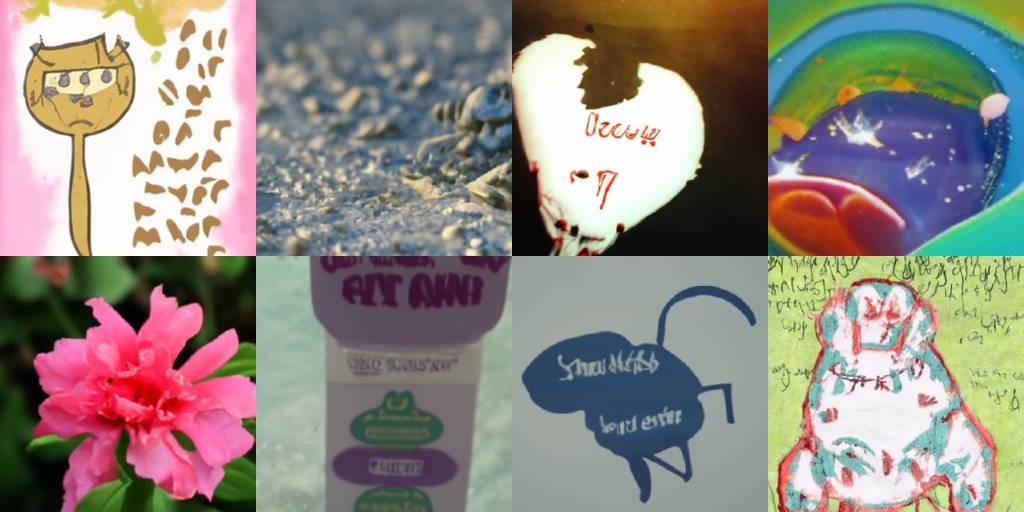}
    }
    \subfigure[\modelname{} (filtered) + unfiltered public CLIP]{
        \includegraphics[width=0.4\textwidth]{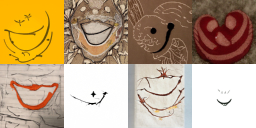}
    }
    \subfigure[Baseline (publicly-available models)]{
        \includegraphics[width=0.4\textwidth]{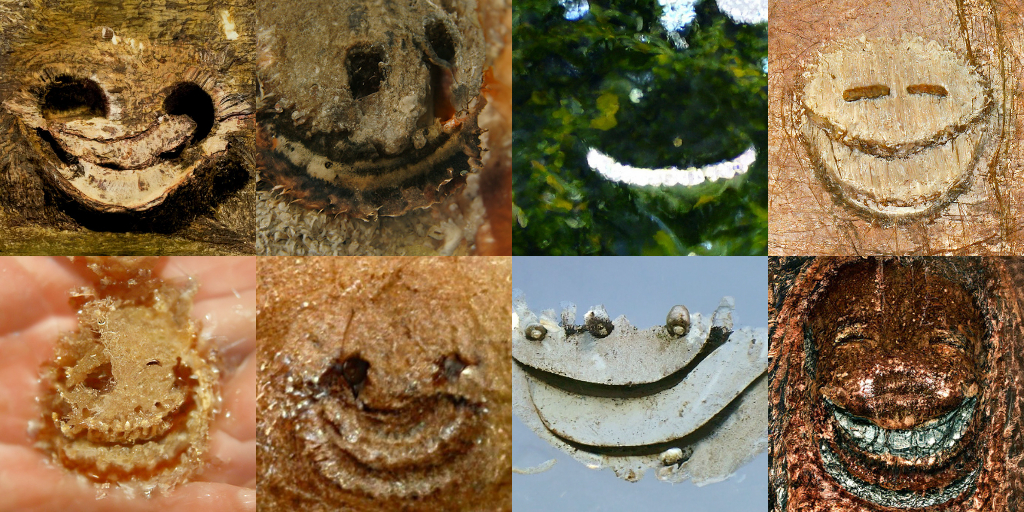}
    }
    \caption{Generations for the prompt ``a happy person''. In (a), we show \modelname{} (filtered) with classifier-free guidance scale 3.0. In (b), we use a publicly-available CLIP model to guide \modelname{} (filtered). In (c), we use a publicly-available CLIP model to guide a publicly-available ImageNet diffusion model.}
    \label{fig:public_happy_person}
    \vskip -0.2in
\end{figure}

\begin{figure}[t]
    \centering
    \subfigure[\modelname{} (filtered)]{
        \includegraphics[width=0.4\textwidth]{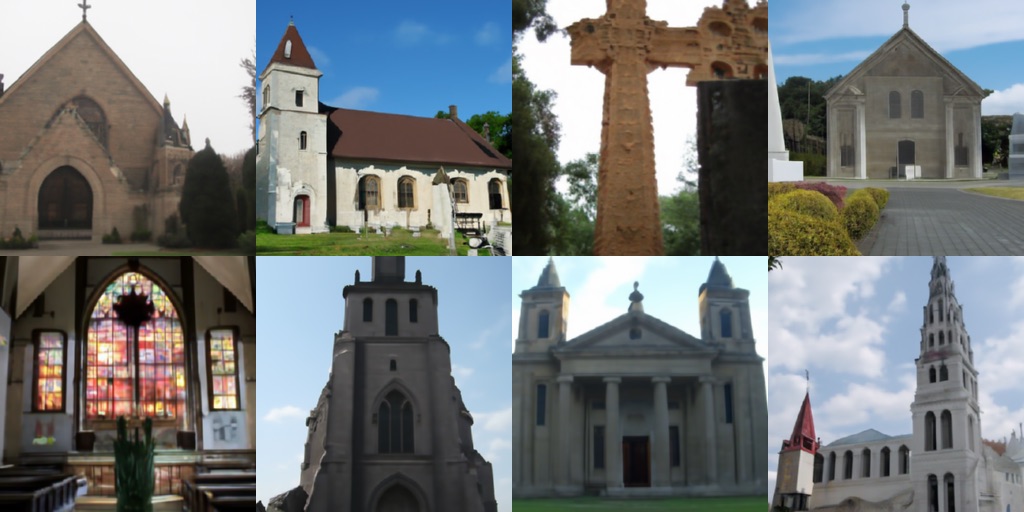}
    }
    \subfigure[\modelname{} (filtered) + unfiltered public CLIP]{
        \includegraphics[width=0.4\textwidth]{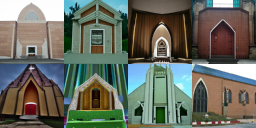}
    }
    \subfigure[Baseline (publicly-available models)]{
        \includegraphics[width=0.4\textwidth]{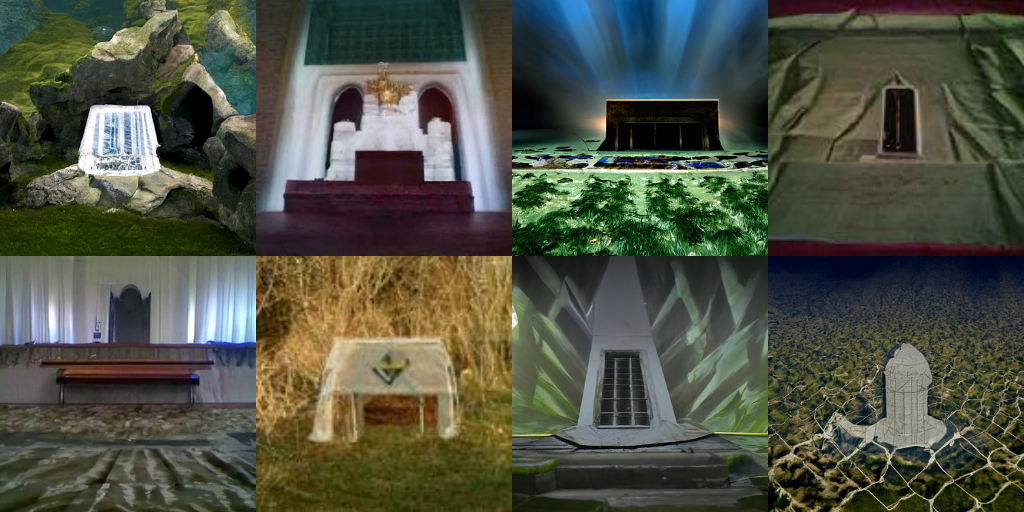}
    }
    \caption{Generations for the prompt ``a place of worship''. In (a), we show \modelname{} (filtered) with classifier-free guidance scale 3.0. In (b), we use a publicly-available CLIP model to guide \modelname{} (filtered). In (c), we use a publicly-available CLIP model to guide a publicly-available ImageNet diffusion model.}
    \label{fig:public_worship}
    \vskip -0.2in
\end{figure}

\subsection{Biases and CLIP Guidance for \modelname{} (filtered)}
\label{app:small_model}
GLIDE (filtered) continues to exhibit bias -- a demonstration both of how biases in image datasets extend beyond those found in images of people, and pointing to biases in the choices we made in filtering. For example, the model produces different outputs when asked to generate toys for boys and toys for girls (Figure \ref{fig:toy_genders}). When asked to generate ``a religious place'', the model tends to gravitate towards church-like buildings, and this bias is amplified by classifier-free guidance (Figure \ref{fig:religion_guidance}).

We expect that our hate symbol classifier has a strong American and Western bias, since it was only trained on two prevalent hate symbols in America. As a result, it is likely that the training data retains images depicting hateful symbols we did not actively filter. However, we do find that the filtered model is less able to generate non-hate symbols (Figure \ref{fig:symbol_knowledge}). We hypothesize that this may be a result of the smaller dataset available to \modelname{} (filtered).

We also incorporated \modelname{} (filtered) into a publicly-available CLIP-guided diffusion program \citep{clipdiff}. We found that the resulting combination had some ability to generate face-like objects (e.g. Figure \ref{fig:public_happy_person}). While, the original CLIP-guided diffusion program using a publicly-available diffusion model often produced more recognizable images in response to our prompts, these findings highlight one of the limitations of our filtering approach. We also found that \modelname{} (filtered) still exhibits a strong Western bias in some cases, often exceeding the bias exhibited by the existing publicly-available diffusion model (Figure \ref{fig:public_worship}).

\section{Additional Samples}
\label{app:add_samples}

\begin{figure*}[t]
    \centering
    \subfigure[DALL-E (Temp 0.85, CLIP reranked top 16 out of 512)]{
        \includegraphics[width=0.45\textwidth]{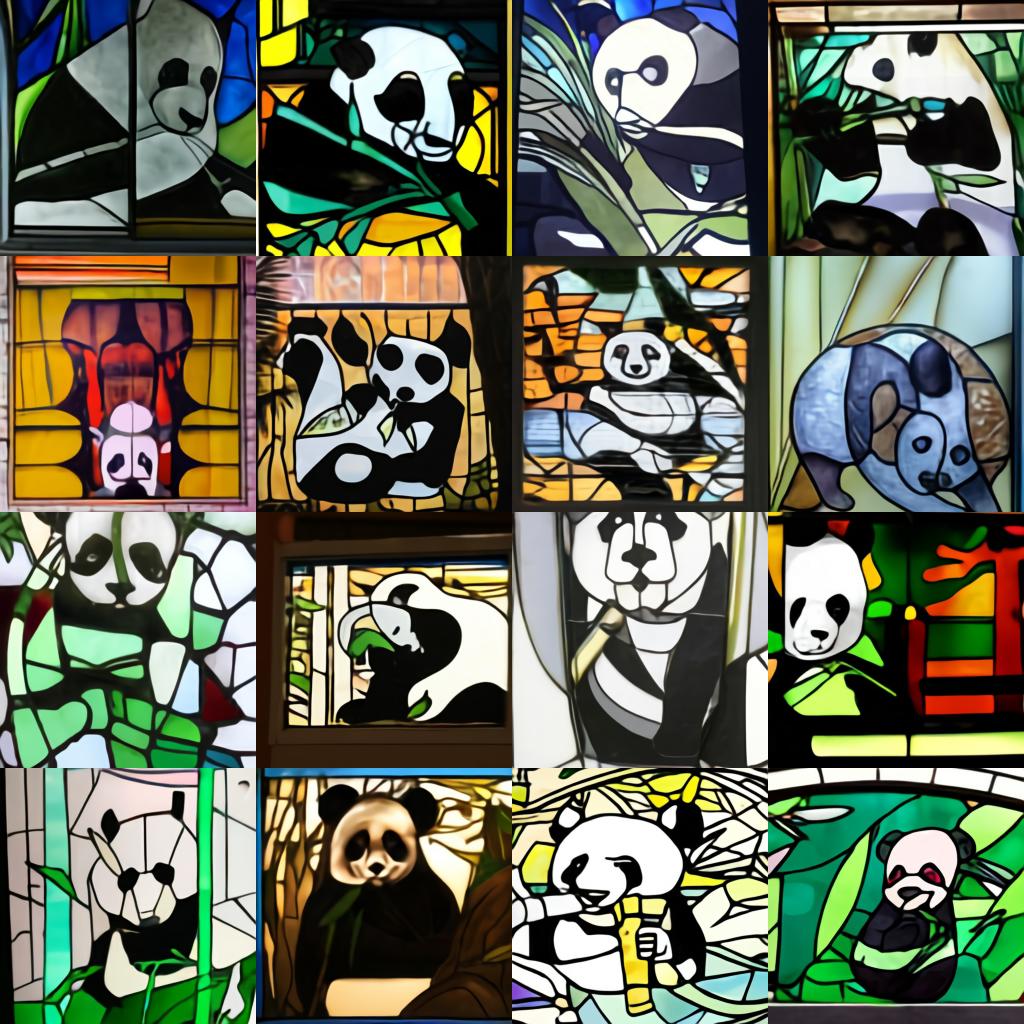}
    }
    \subfigure[\modelname{} (Unguided)]{
        \includegraphics[width=0.45\textwidth]{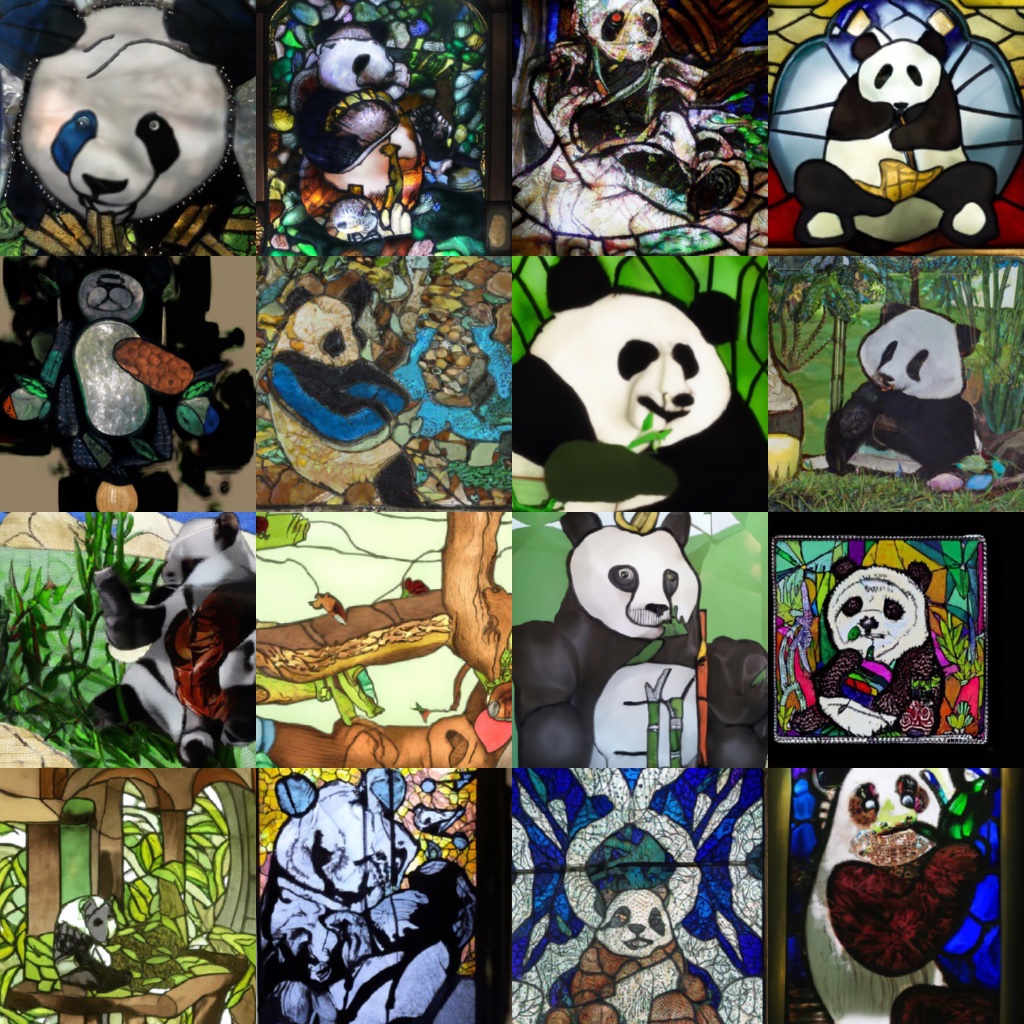}
    }
    \subfigure[\modelname{} (CLIP guidance, scale 2.0)]{
        \includegraphics[width=0.45\textwidth]{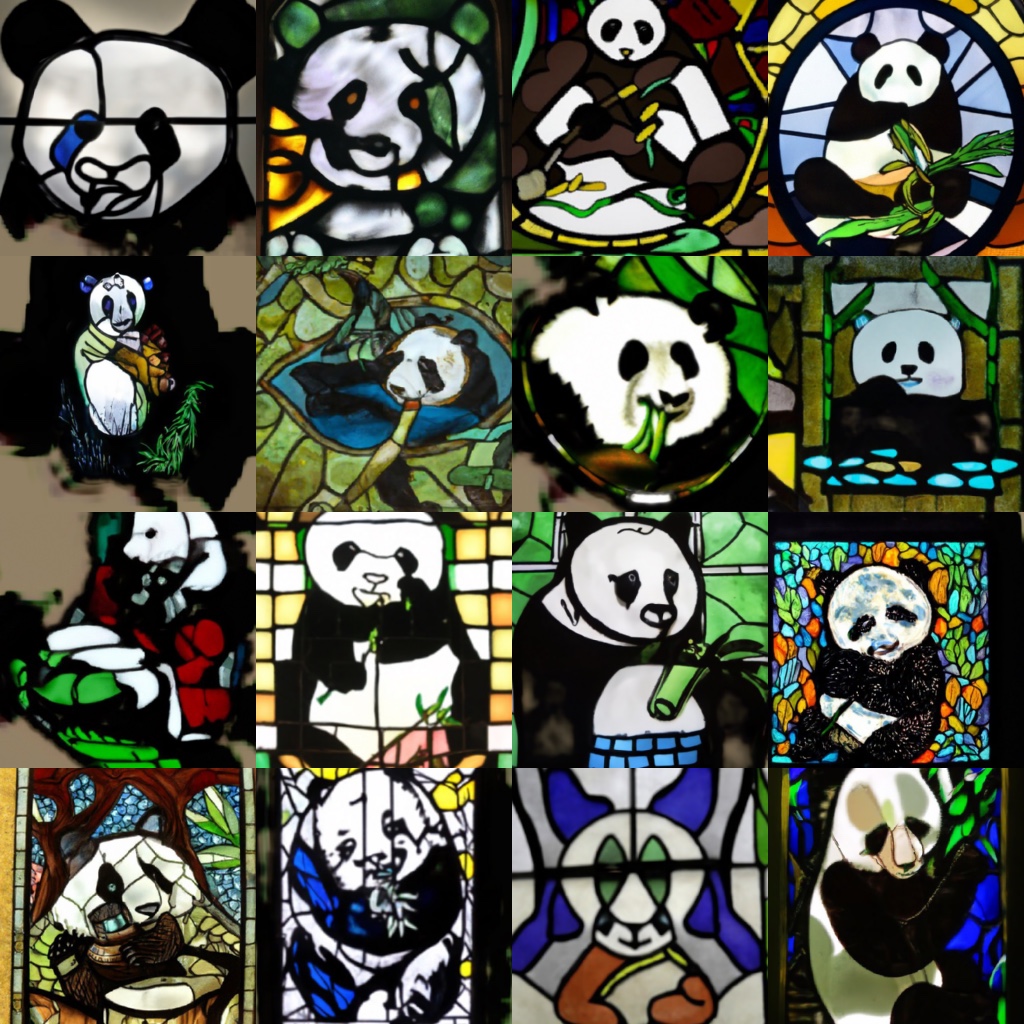}
    }
    \subfigure[\modelname{} (Classifier-free guidance, scale 3.0)]{
        \includegraphics[width=0.45\textwidth]{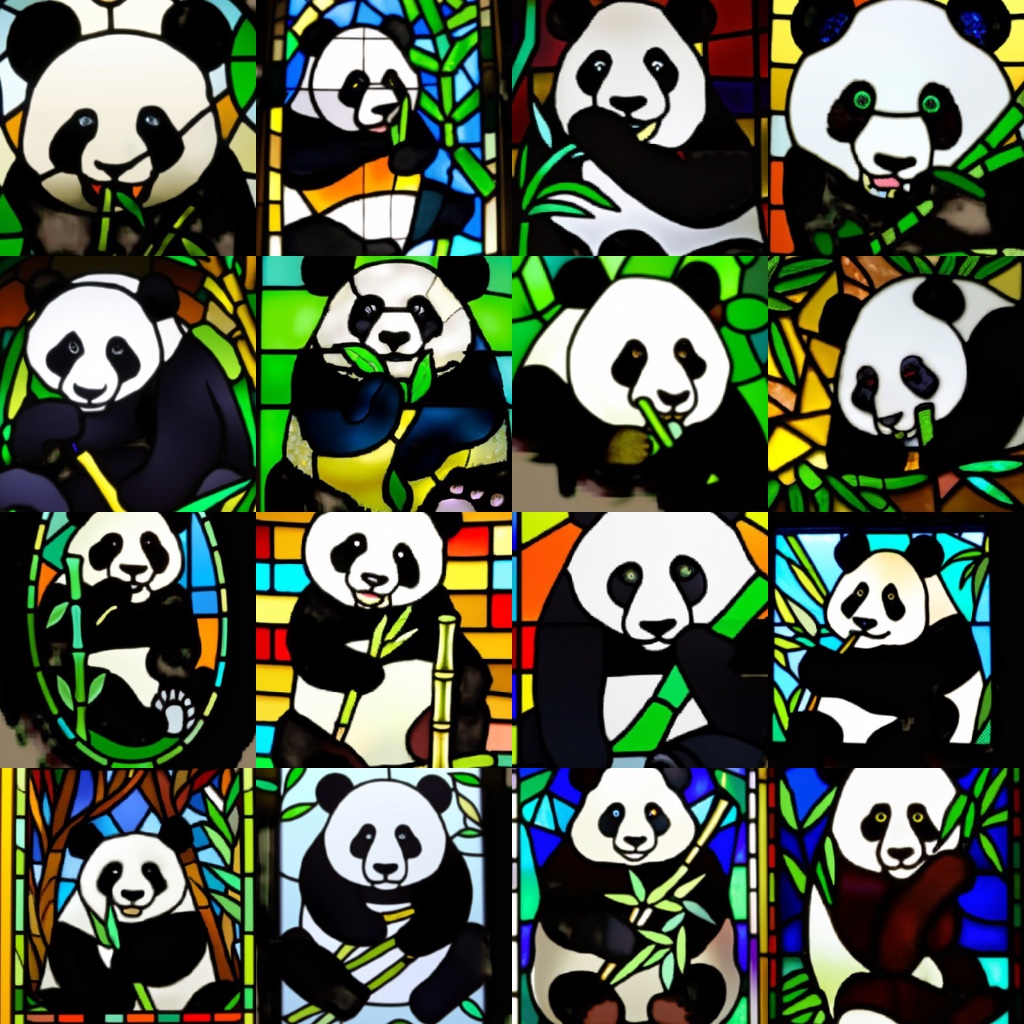}
    }
    \caption{Random samples from DALL-E and \modelname{} on the prompt ``a stained glass window of a panda eating bamboo''.  We do not perform any CLIP reranking for \modelname{}.}
    \label{fig:guidance_comp_bamboo}
\end{figure*}

\begin{figure*}[t]
    \centering
    \subfigure[DALL-E (Temp 0.85, CLIP reranked top 16 out of 512)]{
        \includegraphics[width=0.45\textwidth]{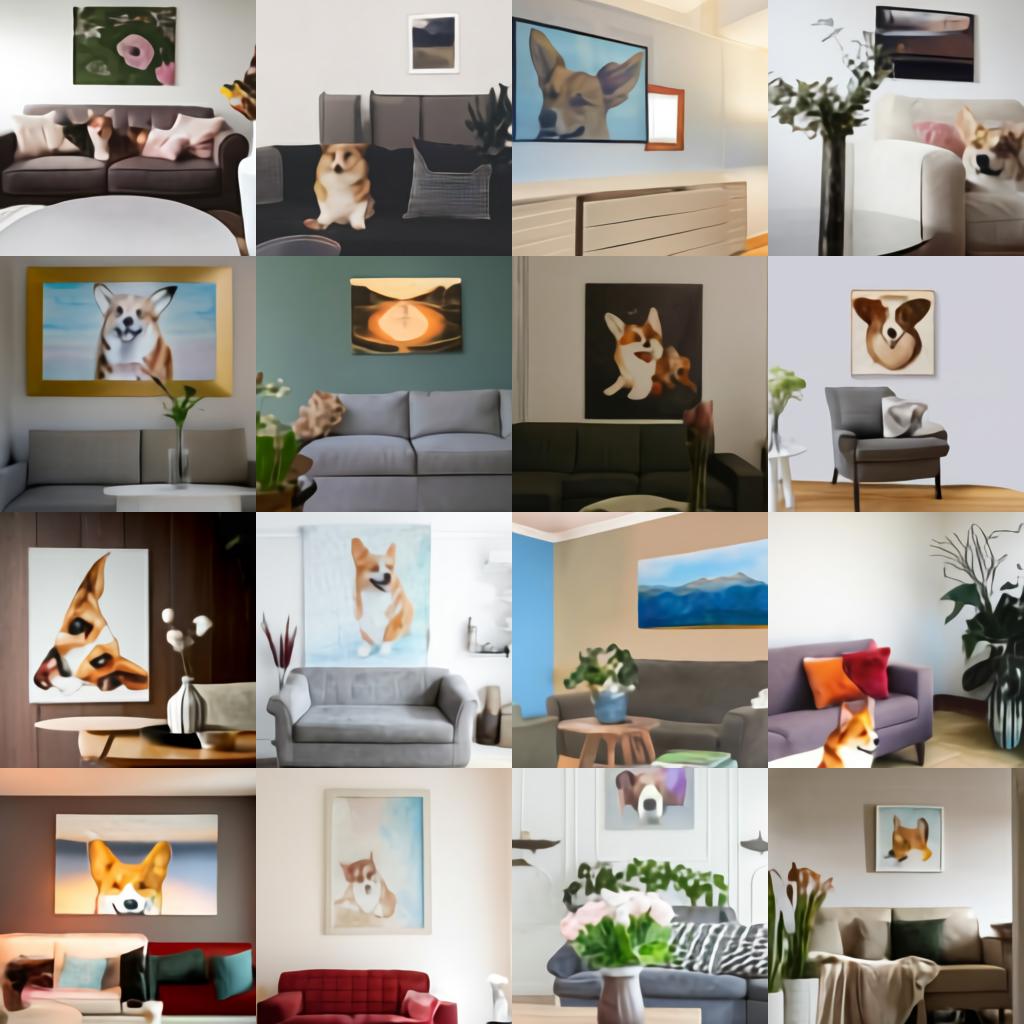}
    }
    \subfigure[\modelname{} (Unguided)]{
        \includegraphics[width=0.45\textwidth]{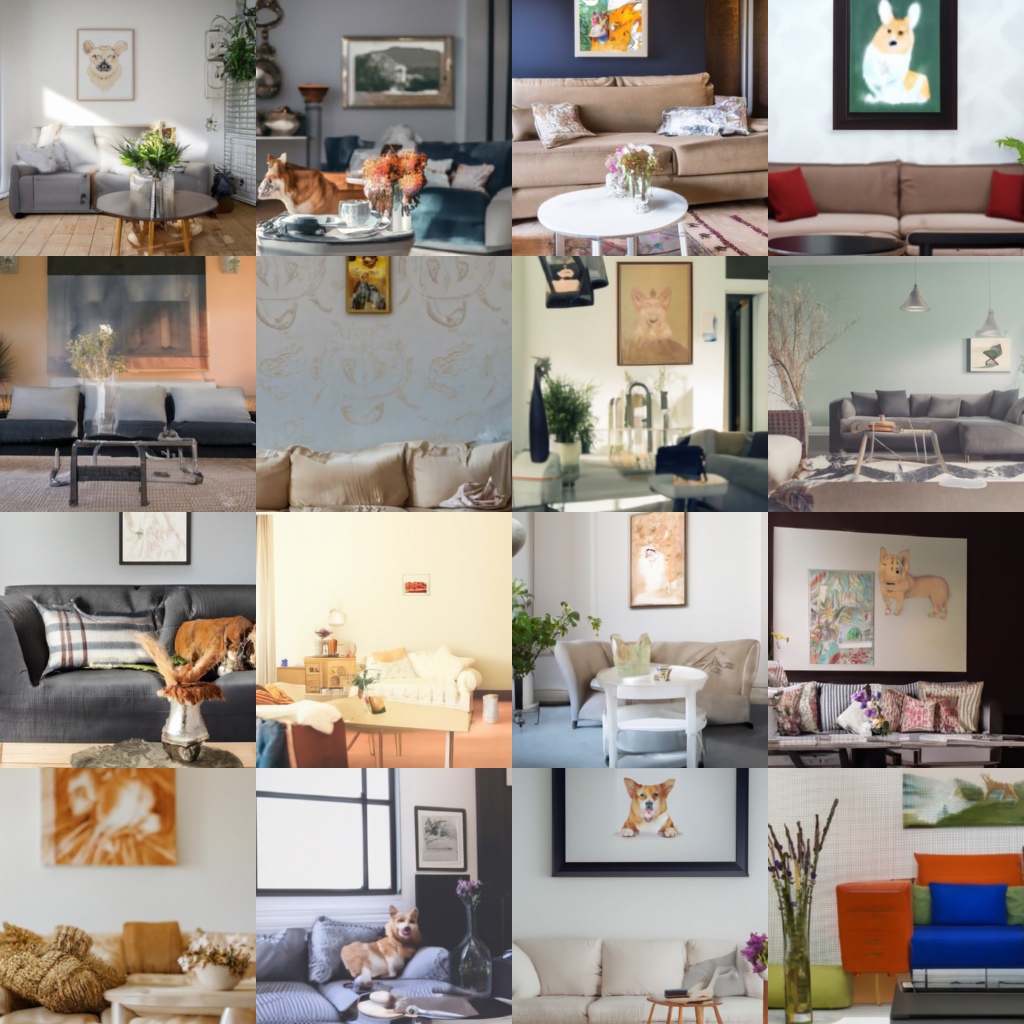}
    }
    \subfigure[\modelname{} (CLIP guidance, scale 2.0)]{
        \includegraphics[width=0.45\textwidth]{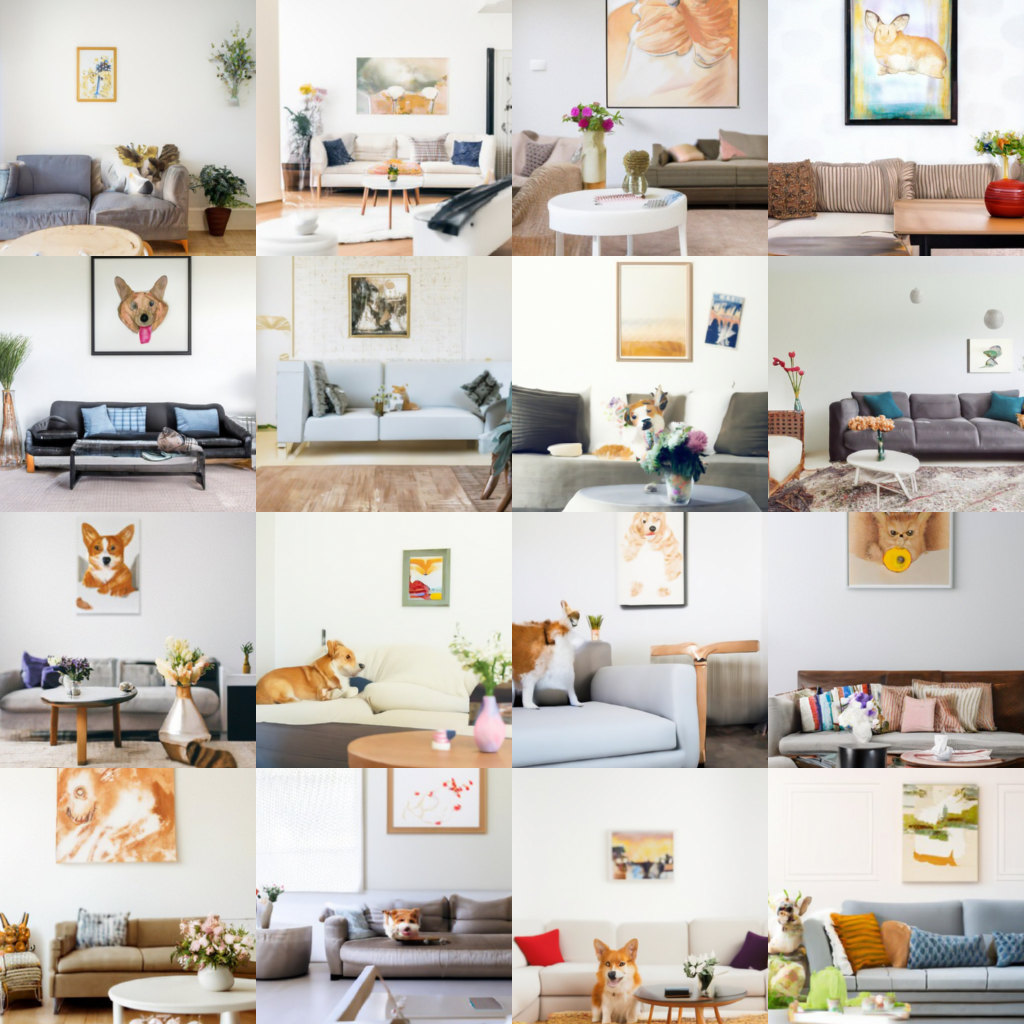}
    }
    \subfigure[\modelname{} (Classifier-free guidance, scale 3.0)]{
        \includegraphics[width=0.45\textwidth]{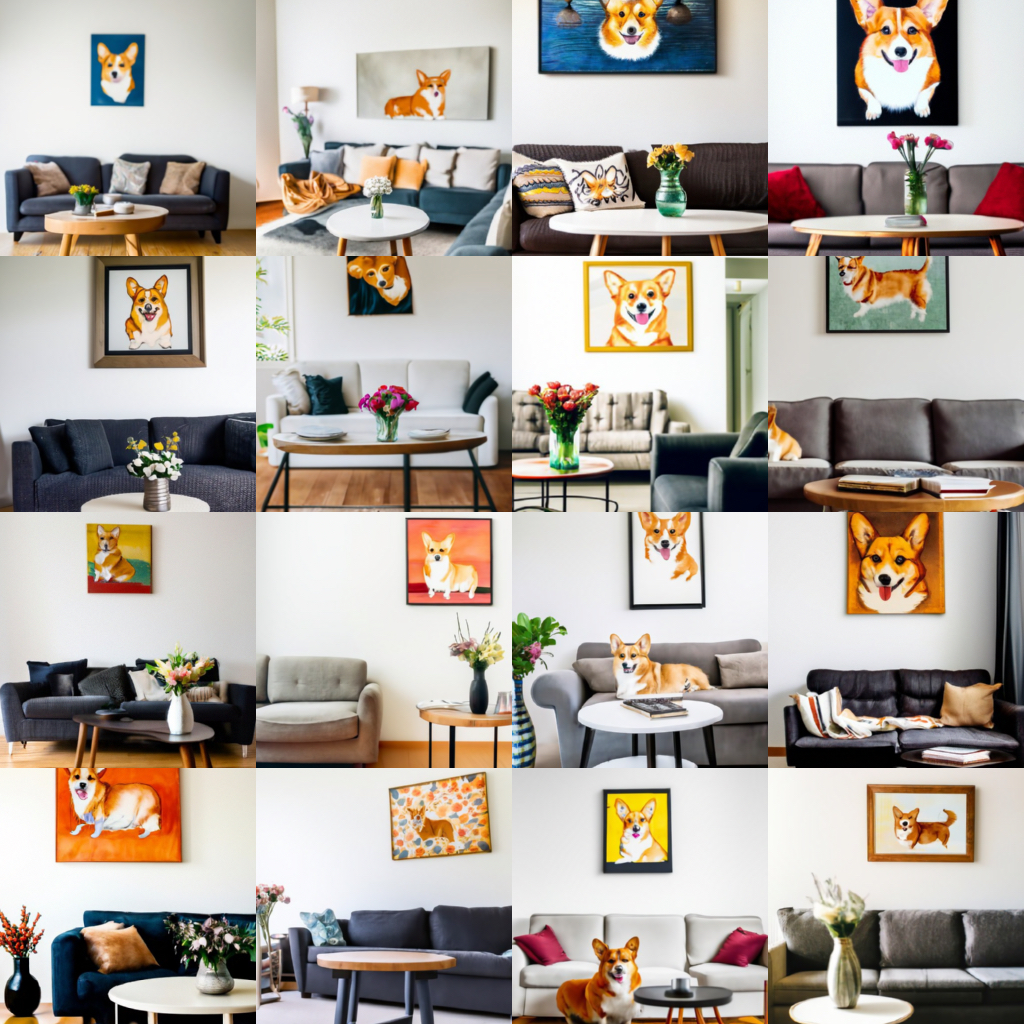}
    }
    \caption{Random samples from DALL-E and \modelname{} on the prompt ``A cozy living room with a painting of a corgi on the wall above a couch and a round coffee table in front of a couch and a vase of flowers on a coffee table''. We do not perform any CLIP reranking for \modelname{}.}
    \label{fig:guidance_comp_living}
\end{figure*}

In Figures \ref{fig:guidance_comp_bamboo} and \ref{fig:guidance_comp_living} we show $4 \times 4$ grids of random samples from our model with no guidance, classifier-free guidance, and CLIP guidance using the same random seeds, as well as samples from DALL-E. We find that classifier-free guidance produces the highest-quality images most reliably. For DALL-E, we sample 512 images for each prompt and select the top 16 using CLIP reranking. For all other sample grids, we show 16 random samples without CLIP reranking.

\end{document}